\PassOptionsToPackage{dvipsnames}{xcolor}
\documentclass[sigconf,nonacm]{aamas} 

\pdfoutput=1
\makeatletter
\@ifclasswith{aamas}{nonacm}{
}{
    \includeonly{1-paper}
}
\makeatother

\makeatletter
\gdef\@copyrightpermission{
  \begin{minipage}{0.2\columnwidth}
   \href{https://creativecommons.org/licenses/by/4.0/}{\includegraphics[width=0.90\textwidth]{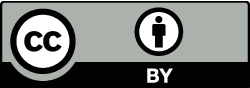}}
  \end{minipage}\hfill
  \begin{minipage}{0.8\columnwidth}
   \href{https://creativecommons.org/licenses/by/4.0/}{This work is licensed under a Creative Commons Attribution International 4.0 License.}
  \end{minipage}
  \vspace{5pt}
}
\makeatother

\setcopyright{ifaamas}
\acmConference[AAMAS '25]{Proc.\@ of the 24th International Conference
on Autonomous Agents and Multiagent Systems (AAMAS 2025)}{May 19 -- 23, 2025}
{Detroit, Michigan, USA}{Y.~Vorobeychik, S.~Das, A.~Nowé  (eds.)}
\copyrightyear{2025}
\acmYear{2025}
\acmDOI{}
\acmPrice{}
\acmISBN{}

\acmSubmissionID{277}

\title{\FORM{}: Learning Expressive and Transferable First-Order Logic Reward Machines}

\author{Leo Ardon}
\affiliation{
  \institution{Imperial College London}
  \city{London}
  \country{United Kingdom}
}
\email{leo.ardon19@ic.ac.uk}
\authornote{Corresponding author.}
\orcid{0000-0003-4400-7127}

\author{Daniel Furelos-Blanco}
\affiliation{
  \institution{Imperial College London}
  \city{London}
  \country{United Kingdom}
}
\email{d.furelos-blanco18@ic.ac.uk}
\orcid{0000-0001-7461-1910}

\author{Roko Parac}
\affiliation{
  \institution{Imperial College London}
  \city{London}
  \country{United Kingdom}
}
\email{roko.parac18@ic.ac.uk}
\orcid{0009-0009-2916-2410}

\author{Alessandra Russo}
\affiliation{
  \institution{Imperial College London}
  \city{London}
  \country{United Kingdom}
}
\email{a.russo@ic.ac.uk}
\authornote{\update{Sponsored in part by DEVCOM Army Research Lab under W911NF2220243 and EPSRC project EP/Y037421/1.}}
\orcid{0000-0002-3318-8711}

\begin{abstract}

Reward machines (RMs) are an effective approach for addressing non-Markovian rewards in reinforcement learning (RL) through finite-state machines. Traditional RMs, which label edges with propositional logic formulae, inherit the limited expressivity of propositional logic. This limitation hinders the learnability and transferability of RMs since complex tasks will require numerous states and edges. To overcome these challenges, we propose First-Order Reward Machines (\FORM{}s), which use first-order logic to label edges, resulting in more compact and transferable RMs. We introduce a novel method for \emph{learning} \FORM{}s and a multi-agent formulation for \emph{exploiting} them and facilitate their transferability, where multiple agents collaboratively learn policies for a shared \FORM{}. Our experimental results demonstrate the scalability of \FORM{}s with respect to traditional RMs. Specifically, we show that \FORM{}s can be effectively learnt for tasks where traditional RM learning approaches fail. We also show significant improvements in learning speed and task transferability thanks to the multi-agent learning framework and the abstraction provided by the first-order language.

\end{abstract}

\ccsdesc[500]{Theory of computation~Reinforcement learning}
\ccsdesc[500]{Computing methodologies~Inductive logic learning}
\ccsdesc[300]{Computing methodologies~Logic programming and answer set programming}

\keywords{Reward Machine; Reinforcement Learning; First-Order Logic}

\usepackage{balance} 
\usepackage{algorithm}
\usepackage[noend]{algorithmic}

\usepackage{newfloat}
\usepackage{listings}
\DeclareCaptionStyle{ruled}{labelfont=normalfont,labelsep=colon,strut=off} 
\floatstyle{ruled}
\newfloat{listing}{tb}{lst}{}
\floatname{listing}{Listing}
\usepackage{xargs}
\usepackage{xspace}
\usepackage{amsmath}

\usepackage{amssymb}
\usepackage{xcolor}
\usepackage{tikz}
\usepackage{marvosym}
\usepackage[marvosym]{tikzsymbols}
\usepackage{multicol}
\usepackage{xspace}
\usepackage[nounderscore]{syntax}
\usepackage{etoolbox}
\usepackage{subcaption}
\usepackage{adjustbox}
\usepackage[outline]{contour}
\usepackage{framed}
\usepackage{bm}
\usepackage{dsfont}

\usetikzlibrary{automata,positioning,arrows.meta}
\usetikzlibrary{matrix}
\tikzset{%
    in place/.style={
        auto=false,
        fill=white,
        inner sep=2pt,
    },
}

\colorlet{yellow}{Goldenrod}
\colorlet{green}{LimeGreen}
\colorlet{gray}{Gray}
\colorlet{red}{Red}
\colorlet{violet}{Orchid}
\colorlet{blue}{Cerulean}
\colorlet{orange}{Orange}
\newcommand{\FORM} {\texttt{FORM}}

\newcommand{\RM} {\mathcal{RM}}
\newcommand{\uI} {u_0}
\newcommand{\uacc} {u_A}
\newcommand{\urej} {u_R}
\newcommand{\transitionFormulaFunction} {\phi}
\newcommand{\deltaR} {\delta_r}
\newcommand{\deltaF} {\delta_u}

\newcommand{\statesSet} {\mathcal{U}}
\newcommand{\observablesSet} {\mathcal{P}}
\newcommand{\observationsSet} {\mathcal{O}}
\newcommand{\observation} {O}

\newcommand{\firstOrderLanguage}{\mathcal{L}}
\newcommandx{\herbrandBase}[1][1=\firstOrderLanguage] {\mathcal{HB}_{#1}}
\newcommandx{\herbrand} {\mathcal{H}}

\newcommand{\transition}[2]{#1 \rightarrow #2}
\newcommand{\maxDisjuncts}{k}

\newcommand{\domain}{\mathcal{C}}
\newcommand{\predicateSet}{\mathcal{K}}

\newcommand{\quantifierSet}{\{ \forall, \exists \}}

\newcommand{\pred}[1]{\displaystyle\mathtt{#1}}
\newcommandx{\predvar}[3][1=P, 2=i] {\pred{#1}_{#2}(#3)}
\newcommand{\true} {\textsc{true}\xspace}
\newcommand{\false} {\textsc{false}\xspace}
\newcommand{\quantifier}{Q}

\newcommand{\bufferSymbol}{B}
\newcommand{\buffer}[1]{\bufferSymbol_{#1}}

\newcommand{\envStatesSet}{S}
\newcommand{\envActionsSet}{A}
\newcommand{\labellingFunction}{L}

\newcommand{\predicateListingColor}{orange}
\newcommand{\listingColor}{ForestGreen}

\renewcommand{\eqref}[1]{Equation~\ref{#1}}

\newcommand{\defref}[1]{Definition~\ref{#1}}
\newcommand{\algoref}[1]{Algorithm~\ref{#1}}
\newcommand{\algolineref}[1]{\texttt{l.\ref{#1}}}
\newcommand{\secref}[1]{Section~\ref{#1}}
\newcommand{\figref}[1]{Figure~\ref{#1}}
\newcommand{\tblref}[1]{Table~\ref{#1}}

\newcommand{\lstref}[1]{Listing~\ref{#1}}
\newcommand{\exref}[1]{Example~\ref{#1}}

\newtheorem{definition}{Definition}[section]

\newtheorem{example}{Example}[section]

\DeclareMathOperator{\codeif}{\mathtt{:-} }

\newcommand{\update}[1]{#1}

\makeatletter
\@ifclasswith{aamas}{nonacm}{
    \newcommand{\appendixref}[1]{Appendix~\ref{#1}}
}{
    \newcommand{\appendixref}[1]{\cite[Appendix~\ref{#1}]{ardon2024form}}
}
\makeatother
\newcommand\ballSize{1.5pt}
\newcommand\ballText{black}

\newcommand{\rmRunningExamplePropHor} {
\begin{tikzpicture}[shorten >=1pt, node distance=2.5cm, on grid, auto]
   \node[state, scale=1.0] (u_0)   {$\uI$}; 
   \node[state, scale=1.0, yshift=-15pt] (u_1) [above right=of u_0] {$u_1$};
   \node[state, scale=1.0, yshift=15pt] (u_2) [below right=of u_0] {$u_2$};
   \node[state, scale=1.0, yshift=-15pt] (u_3) [above right=of u_2]{$u_3$}; 
   \node[state, scale=1.0] (u_4) [right=of u_3]{$u_4$}; 
   \node[state, accepting, scale=1.0] (u_acc) [right=of u_4]{$\uacc$}; 

    \draw 
    (u_0) edge[bend left=35, auto=left, ->] node[midway, circle, fill=yellow, draw=gray, inner sep=\ballSize, xshift=-2pt, yshift=2pt, font=\small, text=\ballText] {$0$} (u_1)
    (u_0) edge[bend right=35, auto=right, ->] node[midway, circle, draw=gray, fill=yellow, inner sep=\ballSize, xshift=-2pt, yshift=-2pt, font=\small, text=\ballText] {$1$} (u_2);

    \draw 
    (u_2) edge[bend right=35, auto=left, ->] node[midway, below, circle, draw=gray, fill=yellow, inner sep=\ballSize, yshift=-1pt, xshift=5pt, font=\small, text=\ballText] {$0$} (u_3)
    (u_1) edge[bend left=35, auto=right, ->] node[midway, below, circle, draw=gray, fill=yellow, inner sep=\ballSize, yshift=10pt, xshift=8pt, font=\small, text=\ballText] {$1$} (u_3);

    \draw 
    (u_3) edge[bend right=35, auto=left, ->] node[midway, below, circle, draw=gray, fill=blue, inner sep=\ballSize, yshift=-5pt, font=\small, text=\ballText] {$4$} (u_4)
    (u_3) edge[bend left=35, auto=right, ->] node[midway, below, circle, draw=gray, fill=blue, inner sep=\ballSize, yshift=15pt, font=\small, text=\ballText] {$5$} (u_4);
 
    \draw 
    (u_4) edge[->] node[midway, draw=gray, fill=green, inner sep=6pt, yshift=3pt, text=\ballText] {} (u_acc);

\end{tikzpicture}
}

\newcommand{\rmRunningExampleFOLHor} {
\begin{tikzpicture}[shorten >=1pt, node distance=2.5cm, on grid, auto]
   \node[state, scale=1.0] (u_0)   {$\uI$}; 
   \node[state, scale=1.0] (u_1) [right=of u_0] {$u_1$};
   \node[state, scale=1.0] (u_2) [right=of u_1] {$u_2$};
   \node[state, accepting, scale=1.0] (u_acc) [right=of u_2]{$\uacc$}; 

    \draw 
    (u_0) edge[->] node[midway, yshift=2pt, xshift=-8pt, font=\small] {$\forall X$.} node[midway, circle, draw=gray, fill=yellow, inner sep=\ballSize, yshift=3pt, xshift=6pt, font=\small, text=\ballText] {$X$} (u_1);

    \draw 
    (u_1) edge[->] node[midway, yshift=2pt, xshift=-8pt, font=\small] {$\exists X$.} node[midway, circle, draw=gray, fill=blue, inner sep=\ballSize, yshift=3pt, xshift=6pt, font=\small, text=\ballText] {$X$} (u_2);

    \draw 
    (u_2) edge[->] node[midway, draw=gray, fill=green, inner sep=6pt, yshift=3pt, text=\ballText] {} (u_acc);

\end{tikzpicture}
}

\newcommand{\rmExistentialMotivationProp} {
\begin{tikzpicture}[shorten >=1pt, node distance=1.5cm, on grid, auto]
   \node[state, scale=0.75] (u_1)   {$u_1$}; 
   \node[state, scale=0.75] (u_2) [right=of u_1] {$u_2$};

    \draw 
    (u_1) edge[bend left=35, auto=left, ->] node[midway, circle, draw=blue, fill=blue, inner sep=\ballSize, yshift=2pt, font=\scriptsize, text=\ballText] {$5$} (u_2)
    (u_1) edge[bend right=35, auto=right, ->] node[midway, circle, draw=blue, fill=blue, inner sep=\ballSize, yshift=-2pt, font=\scriptsize, text=\ballText] {$4$} (u_2);
\end{tikzpicture}
}

\newcommand{\rmExistentialMotivationFOL} {
\begin{tikzpicture}[shorten >=1pt, node distance=1.5cm, on grid, auto]
   \node[state, scale=0.75] (u_1)   {$u_1$}; 
   \node[state, scale=0.75] (u_2) [right=of u_1] {$u_2$};

    \draw 
    (u_1) edge[->] node[midway, yshift=1pt, xshift=-6pt, font=\scriptsize] {$\exists X$.} node[midway, circle, draw=gray, fill=blue, inner sep=\ballSize, yshift=2pt, xshift=6pt, font=\scriptsize, text=\ballText] {$X$} (u_2);

\end{tikzpicture}
}

\newcommand{\rmUniversalMotivationProp} {
\begin{tikzpicture}[shorten >=1pt, node distance=1.5cm, on grid, auto]
   \def\c{yellow}
   
   \node[state, scale=0.75] (u_0)   {$\uI$}; 
   \node[state, scale=0.75, yshift=-15pt] (u_1) [above right=of u_0] {$u_1$};
   \node[state, scale=0.75, yshift=15pt] (u_2) [below right=of u_0] {$u_2$};
   \node[state, scale=0.75, yshift=15pt] (u_3) [below right=of u_1]{$u_3$}; 

    \draw 
    (u_0) edge[->] node[midway, above, circle, draw=gray, fill=\c, inner sep=\ballSize, xshift=-5pt, yshift=2pt, font=\scriptsize, text=\ballText] {$1$} (u_1);

    \draw 
    (u_1) edge[->] node[midway, above, circle, draw=gray, fill=\c, inner sep=\ballSize, xshift=5pt, yshift=2pt, font=\scriptsize, text=\ballText] {$2$} (u_3);

    \draw 
    (u_0) edge[->] node[midway, below, circle, draw=gray, fill=\c, inner sep=\ballSize, xshift=-5pt, yshift=-2pt, font=\scriptsize, text=\ballText] {$2$} (u_2);

    \draw 
    (u_2) edge[->] node[midway, below, circle, draw=gray, fill=\c, inner sep=\ballSize, xshift=5pt, yshift=-2pt, font=\scriptsize, text=\ballText] {$1$} (u_3);
    
\end{tikzpicture}
}

\newcommand{\rmUniversalMotivationFOL} {
\begin{tikzpicture}[shorten >=1pt, node distance=1.5cm, on grid, auto]
    \def\c{yellow}

   \node[state, scale=0.75] (u_0)   {$\uI$}; 
   \node[state, scale=0.75] (u_1) [right=of u_0]{$u_1$}; 

   \draw 
    (u_0) edge[->] node[midway, yshift=1.5pt, xshift=-6pt, font=\scriptsize] {$\forall X$.} node[midway, circle, draw=gray, fill=\c, inner sep=\ballSize, yshift=2pt, xshift=6pt, font=\scriptsize, text=\ballText] {$X$} (u_1);

\end{tikzpicture}
}

\newcommand{\motivationFig}[1]{
\begin{figure}[#1]
    \centering
    \captionsetup{justification=centering}
    \captionsetup[subfigure]{justification=centering}

    \begin{subfigure}[b]{0.4\textwidth}
        \begin{subfigure}[b]{0.4\textwidth}
            \centering
            \rmExistentialMotivationProp
        \end{subfigure}
        \hfill
        \raisebox{20pt}{
            \scalebox{1.5}{
                {$\Rightarrow$}%
            }
        }
        \begin{subfigure}[b]{0.4\textwidth}
            \centering
            \raisebox{14pt}{
                \rmExistentialMotivationFOL
            }
        \end{subfigure}
        \Description{Demonstration of how existentially quantified sentences can reduce the number of edges required to encode a task compared to propositional RM.}
        \caption{Existentially quantified formulae.}
        \label{fig:existential_motivation}
    \end{subfigure}

    \begin{subfigure}[b]{0.4\textwidth}
        \begin{subfigure}[b]{0.4\textwidth}
            \centering
            \rmUniversalMotivationProp
        \end{subfigure}
        \hfill
        \raisebox{25pt}{
            \scalebox{1.5}{
                {$\Rightarrow$}%
            }
        }
        \begin{subfigure}[b]{0.4\textwidth}
            \centering
            \raisebox{20pt}{
                \rmUniversalMotivationFOL
            }
        \end{subfigure}
        \caption{Universally quantified formulae.}
        \Description{Demonstration of how universally quantified sentences can reduce the number of states required to encode a task compared to propositional RM.}
        \label{fig:universal_motivation}
    \end{subfigure}
    \caption{Equivalence between Propositional RMs and \FORM{}s.}
    \Description{Demonstration \FORM{}s reduce the number of nodes and edges required to encode the task.}
\end{figure}
}

\newcommand{\rmUniversalMotivationPropThree} {
\begin{tikzpicture}[shorten >=1pt, node distance=2cm, on grid, auto]
    \def\c{yellow}
    
    \node[state, scale=0.75] (u_0)   {$u_0$};
    \node[state, scale=0.75] (u_1) [above right=of u_0] {$u_1$};
    \node[state, scale=0.75] (u_2) [below right=of u_0] {$u_2$};
    \node[state, scale=0.75] (u_3) [right=of u_0] {$u_3$};
    \node[state, scale=0.75] (u_4) [right=of u_1] {$u_4$};
    \node[state, scale=0.75] (u_5) [right=of u_3] {$u_5$};
    \node[state, scale=0.75] (u_6) [right=of u_2] {$u_6$};
    \node[state, scale=0.75] (u_7) [right=of u_5] {$u_7$};

    \draw 
    (u_0) edge[->] node[midway, above, circle, draw=gray, fill=\c, inner sep=\ballSize, xshift=-6pt, yshift=-1pt, font=\scriptsize, text=\ballText] {$1$} (u_1);
    \draw 
    (u_0) edge[->] node[midway, above, circle, draw=gray, fill=\c, inner sep=\ballSize, xshift=-6pt, yshift=-8pt, font=\scriptsize, text=\ballText] {$2$} (u_2);
    \draw 
    (u_0) edge[->] node[midway, above, circle, draw=gray, fill=\c, inner sep=\ballSize, xshift=0pt, yshift=2pt, font=\scriptsize, text=\ballText] {$3$} (u_3);

    \draw 
    (u_1) edge[->] node[midway, above, circle, draw=gray, fill=\c, inner sep=\ballSize, xshift=0pt, yshift=2pt, font=\scriptsize, text=\ballText] {$2$} (u_4);
    \draw 
    (u_1) edge[->] node[midway, above, circle, draw=gray, fill=\c, inner sep=\ballSize, xshift=4pt, yshift=0pt, font=\scriptsize, text=\ballText] {$3$} (u_5);

    \draw 
    (u_3) edge[->] node[midway, above, circle, draw=gray, fill=\c, inner sep=\ballSize, xshift=-1pt, yshift=2pt, font=\scriptsize, text=\ballText] {$1$} (u_5);

    \draw 
    (u_5) edge[->] node[midway, above, circle, draw=gray, fill=\c, inner sep=\ballSize, xshift=-3pt, yshift=2pt, font=\scriptsize, text=\ballText] {$2$} (u_7);

    \draw 
    (u_2) edge[->] node[midway, above, circle, draw=gray, fill=\c, inner sep=\ballSize, xshift=-1pt, yshift=2pt, font=\scriptsize, text=\ballText] {$3$} (u_6);

    \draw 
    (u_3) edge[->] node[midway, above, circle, draw=gray, fill=\c, inner sep=\ballSize, xshift=5pt, yshift=-2pt, font=\scriptsize, text=\ballText] {$2$} (u_6);

    \draw 
    (u_6) edge[->] node[midway, above, circle, draw=gray, fill=\c, inner sep=\ballSize, xshift=-2pt, yshift=-14pt, font=\scriptsize, text=\ballText] {$1$} (u_7);

    \draw 
    (u_4) edge[->] node[midway, above, circle, draw=gray, fill=\c, inner sep=\ballSize, xshift=0pt, yshift=4pt, font=\scriptsize, text=\ballText] {$3$} (u_7);

    \draw 
    (u_2) edge[bend left=45, ->] node[midway, above, circle, draw=gray, fill=\c, inner sep=\ballSize, xshift=0pt, yshift=4pt, font=\scriptsize, text=\ballText] {$1$} (u_4);

\end{tikzpicture}
}

\newcommand{\universalMotivationFigThree}[1] {
\begin{figure}[#1]
    \centering
    \captionsetup{justification=centering}
    \captionsetup[subfigure]{justification=centering}
    \begin{subfigure}[b]{0.45\textwidth}
        \centering
        \raisebox{5pt}{
            \rmUniversalMotivationPropThree
        }
    \end{subfigure}
    \hfill
    \raisebox{10pt}{
        \scalebox{1.5}{
            {$\Downarrow$}%
        }
    }
    \hfill
    \begin{subfigure}[b]{0.45\textwidth}
        \centering
        \raisebox{10pt}{
            \rmUniversalMotivationFOL
        }
    \end{subfigure}
    \caption{Equivalence between a propositional and a first-order RM for the \textsc{All-Yellow} task with three yellow checkpoints .}
    \Description{Demonstration of how universally quantified sentences can reduce the number of states required to encode a task compared to propositional RM.}
    \label{fig:universal_motivation_3}
\end{figure}
}

\newcommand{\rmAllYellowFOLHor} {
\begin{tikzpicture}[shorten >=1pt, node distance=1.5cm, on grid, auto]
   \node[state, scale=0.75] (u_0)   {$\uI$}; 
   \node[state, scale=0.75] (u_1) [right=of u_0] {$u_1$};
   \node[state, accepting, scale=0.75,] (u_acc) [right=of u_1]{$\uacc$}; 

    \draw 
    (u_0) edge[->] node[midway, yshift=2pt, xshift=-8pt, font=\scriptsize] {$\forall X$.} node[midway, circle, draw=gray, fill=yellow, inner sep=\ballSize, yshift=3.5pt, xshift=4pt, font=\scriptsize, text=\ballText] {$X$} (u_1);

    \draw 
    (u_1) edge[->] node[midway, draw=gray, fill=green, inner sep=5pt, yshift=3pt, xshift=-1pt, text=\ballText] {} (u_acc);

\end{tikzpicture}
}

\newcommand{\rmAllYellowPropHor} {
\begin{tikzpicture}[shorten >=1pt, node distance=1.5cm, on grid, auto]
   \def\c{yellow}
   
   \node[state, scale=0.75] (u_0)   {$\uI$}; 
   \node[state, scale=0.75, yshift=-15pt] (u_1) [above right=of u_0] {$u_1$};
   \node[state, scale=0.75, yshift=15pt] (u_2) [below right=of u_0] {$u_2$};
   \node[state, scale=0.75, yshift=15pt] (u_3) [below right=of u_1]{$u_3$};
   \node[state, accepting, scale=0.75] (u_acc) [right=of u_3]{$\uacc$};

    \draw 
    (u_0) edge[->] node[midway, above, circle, draw=gray, fill=\c, inner sep=\ballSize, xshift=-5pt, yshift=2pt, font=\scriptsize, text=\ballText] {$1$} (u_1);

    \draw 
    (u_1) edge[->] node[midway, above, circle, draw=gray, fill=\c, inner sep=\ballSize, xshift=5pt, yshift=2pt, font=\scriptsize, text=\ballText] {$2$} (u_3);

    \draw 
    (u_0) edge[->] node[midway, below, circle, draw=gray, fill=\c, inner sep=\ballSize, xshift=-5pt, yshift=-2pt, font=\scriptsize, text=\ballText] {$2$} (u_2);

    \draw 
    (u_2) edge[->] node[midway, below, circle, draw=gray, fill=\c, inner sep=\ballSize, xshift=5pt, yshift=-2pt, font=\scriptsize, text=\ballText] {$1$} (u_3);

    \draw 
    (u_3) edge[->] node[midway, draw=gray, fill=green, inner sep=5pt, yshift=3pt, xshift=-1pt, text=\ballText] {} (u_acc);
    
\end{tikzpicture}
}

\newcommand{\rmGreenNoLavaFOLHor} {
\begin{tikzpicture}[shorten >=1pt, node distance=1.5cm, on grid, auto]
   \node[state, scale=0.75] (u_0)   {$\uI$}; 
   \node[state, scale=0.75] (u_1) [right=of u_0] {$u_1$};
   \node[state, accepting, scale=0.75,] (u_acc) [right=of u_1]{$\uacc$}; 
   \node[state, accepting, scale=0.75, yshift=15pt] (u_rej) [below=of u_1]{$\urej$}; 

    \draw 
    (u_0) edge[->] node[midway, yshift=8.5pt, xshift=-20pt, font=\scriptsize] {$\exists X$.} node[midway, circle, draw=gray, fill=green, inner sep=\ballSize, yshift=9pt, xshift=-8pt, font=\scriptsize, text=\ballText] {$X$} node[midway, yshift=8pt, xshift=3pt, font=\scriptsize] {$\bigwedge \neg$} node[midway, circle, draw=gray, fill=green, inner sep=\ballSize, yshift=9pt, xshift=15pt, font=\scriptsize, text=\ballText] {$12$} (u_1);

    \draw 
    (u_0) edge[->] node[midway, draw=gray, fill=orange, inner sep=5pt, yshift=-12pt, xshift=-12pt, text=\ballText] {} (u_rej);
    
    \draw 
    (u_1) edge[->] node[midway, draw=gray, fill=orange, inner sep=5pt, yshift=1pt, xshift=4pt, text=\ballText] {} (u_rej);

    \draw 
    (u_1) edge[->] node[midway, draw=gray, fill=green, inner sep=5pt, yshift=9pt, xshift=-1pt, text=\ballText] {} (u_acc);

\end{tikzpicture}
}

\newcommand{\rmGreenNoLavaPropHor} {
\begin{tikzpicture}[shorten >=1pt, node distance=1.5cm, on grid, auto]
   \node[state, scale=0.75] (u_0)   {$\uI$}; 
   \node[state, scale=0.75] (u_1) [right=of u_0] {$u_1$};
   \node[state, accepting, scale=0.75,] (u_acc) [right=of u_1]{$\uacc$}; 
   \node[state, accepting, scale=0.75, yshift=15pt] (u_rej) [below=of u_1]{$\urej$}; 
    
    \draw 
    (u_0) edge[bend left=90, auto=left, ->] node[midway, circle, draw=gray, fill=green, inner sep=\ballSize, yshift=3.5pt, xshift=-1pt, font=\scriptsize, text=\ballText] {$10$} (u_1);

    \draw 
    (u_0) edge[->] node[midway, circle, draw=gray, fill=green, inner sep=\ballSize, yshift=3.5pt, xshift=-1pt, font=\scriptsize, text=\ballText] {$11$} (u_1);

    \draw 
    (u_0) edge[->] node[midway, draw=gray, fill=orange, inner sep=5pt, yshift=-12pt, xshift=-12pt, text=\ballText] {} (u_rej);
    
    \draw 
    (u_1) edge[->] node[midway, draw=gray, fill=orange, inner sep=5pt, yshift=1pt, xshift=4pt, text=\ballText] {} (u_rej);

    \draw 
    (u_1) edge[->] node[midway, draw=gray, fill=green, inner sep=5pt, yshift=9pt, xshift=-1pt, text=\ballText] {} (u_acc);

\end{tikzpicture}
}

\newcommand{\rmBlueAllYellowSevenFOLHor} {
\begin{tikzpicture}[shorten >=1pt, node distance=1.5cm, on grid, auto]
   \node[state, scale=0.75] (u_0)   {$\uI$}; 
   \node[state, scale=0.75] (u_1) [right=of u_0] {$u_1$};
   \node[state, scale=0.75] (u_2) [right=of u_1] {$u_2$};
   \node[state, scale=0.75] (u_3) [right=of u_2] {$u_3$};
   \node[state, accepting, scale=0.75,] (u_acc) [right=of u_3]{$\uacc$}; 

    \draw 
    (u_0) edge[->] node[midway, yshift=2pt, xshift=-8pt, font=\scriptsize] {$\exists X$.} node[midway, circle, draw=gray, fill=blue, inner sep=\ballSize, yshift=3.5pt, xshift=4pt, font=\scriptsize, text=\ballText] {$X$} (u_1);

    \draw 
    (u_1) edge[->] node[midway, yshift=2pt, xshift=-8pt, font=\scriptsize] {$\forall X$.} node[midway, circle, draw=gray, fill=yellow, inner sep=\ballSize, yshift=3.5pt, xshift=4pt, font=\scriptsize, text=\ballText] {$X$} (u_2);

    \draw 
    (u_2) edge[->] node[midway, circle, draw=gray, fill=violet, inner sep=\ballSize, yshift=3.5pt, xshift=-1pt, font=\scriptsize, text=\ballText] {$7$} (u_3);

    \draw 
    (u_3) edge[->] node[midway, draw=gray, fill=green, inner sep=5pt, yshift=3pt, xshift=-1pt, text=\ballText] {} (u_acc);

\end{tikzpicture}
}

\newcommand{\rmBlueAllYellowSevenPropHor} {
\begin{tikzpicture}[shorten >=1pt, node distance=1.5cm, on grid, auto]
   \node[state, scale=0.75] (u_0)   {$\uI$}; 
   \node[state, scale=0.75] (u_1) [right=of u_0]  {$u_1$}; 
   \node[state, scale=0.75, yshift=-15pt] (u_2) [above right=of u_1] {$u_2$};
   \node[state, scale=0.75, yshift=15pt] (u_3) [below right=of u_1] {$u_3$};
   \node[state, scale=0.75, yshift=15pt] (u_4) [below right=of u_2]{$u_4$}; 
   \node[state, scale=0.75] (u_5) [right=of u_4]{$u_5$}; 
   \node[state, accepting, scale=0.75,] (u_acc) [right=of u_5]{$\uacc$}; 

    \draw 
    (u_0) edge[bend left=35, auto=left, ->] node[midway, circle, draw=blue, fill=blue, inner sep=\ballSize, yshift=2pt, font=\scriptsize, text=\ballText] {$5$} (u_1)
    (u_0) edge[bend right=35, auto=right, ->] node[midway, circle, draw=blue, fill=blue, inner sep=\ballSize, yshift=-2pt, font=\scriptsize, text=\ballText] {$4$} (u_1);

    \draw 
    (u_1) edge[->] node[midway, above, circle, draw=gray, fill=yellow, inner sep=\ballSize, xshift=-5pt, yshift=2pt, font=\scriptsize, text=\ballText] {$1$} (u_2);

    \draw 
    (u_2) edge[->] node[midway, above, circle, draw=gray, fill=yellow, inner sep=\ballSize, xshift=5pt, yshift=2pt, font=\scriptsize, text=\ballText] {$0$} (u_4);

    \draw 
    (u_1) edge[->] node[midway, below, circle, draw=gray, fill=yellow, inner sep=\ballSize, xshift=-5pt, yshift=-2pt, font=\scriptsize, text=\ballText] {$0$} (u_3);

    \draw 
    (u_3) edge[->] node[midway, below, circle, draw=gray, fill=yellow, inner sep=\ballSize, xshift=5pt, yshift=-2pt, font=\scriptsize, text=\ballText] {$1$} (u_4);

    \draw 
    (u_4) edge[->] node[midway, above, circle, draw=gray, fill=violet, inner sep=\ballSize, xshift=-1pt, yshift=2pt, font=\scriptsize, text=\ballText] {$7$} (u_5);

    \draw 
    (u_5) edge[->] node[midway, draw=gray, fill=green, inner sep=5pt, yshift=3pt, xshift=-1pt, text=\ballText] {} (u_acc);

\end{tikzpicture}
}
\newcommand{\minigridTwoObj} {

\begin{tikzpicture}

\def\gridsize{13}

\foreach \x in {1,...,\gridsize}
    \foreach \y in {1,...,\gridsize} {
        \fill[black] (\x-1,\y-1) rectangle (\x,\y);
        \draw[gray] (\x-1,\y-1) rectangle (\x,\y);
    }

\foreach \i in {1,...,\gridsize} {
    \fill[gray] (0, \i-1) rectangle (1, \i);
    \fill[gray] (\gridsize-1, \i-1) rectangle (\gridsize, \i);
}

\foreach \i in {1,...,\gridsize} {
    \fill[gray] (\i-1, 0) rectangle (\i, 1);
    \fill[gray] (\i-1, \gridsize-1) rectangle (\i, \gridsize);
}

\def\wallpos{{
    {7,12},{7,11},{7,9},{7,8},{7,7},{7,6},{7,5},{7,4},{7,2},
    {2,7},{4,7},{5,7},{6,7},
    {8,6},{9,6},{11,6},{12,6},
}}

\foreach \i in {0,...,16} {
    \fill[gray] (\wallpos[\i][0]-1, \wallpos[\i][1]-1) rectangle (\wallpos[\i][0], \wallpos[\i][1]);
}

\def\goalpos{{12,2}}
\fill[green] (\goalpos[0]-1, \goalpos[1]-1) rectangle (\goalpos[0], \goalpos[1]);

\def\agentpos{{3,11}}
\draw[fill=red] (\agentpos[0]-0.5,\agentpos[1]-0.1) -- (\agentpos[0]-0.8,\agentpos[1]-0.9) -- (\agentpos[0]-0.2,\agentpos[1]-0.9) -- cycle;

\def\objpos{8/8/yellow, 11/5/yellow, 9/10/red, 4/4/red, 3/12/blue, 4/10/blue, 6/12/violet, 3/9/violet, 2/11/lightgray, 11/11/lightgray, 8/7/green, 10/8/green}
\foreach \x/\y/\col [count=\k from 0] in \objpos
{
    \fill[color=\col] (\x - 0.5, \y - 0.5) circle (0.33);
    \node[text=black] at (\x - 0.5, \y - 0.5) {$\k$};
}
\end{tikzpicture}
}

\DeclareRobustCommand{\goalCell}{
\trimbox{1pt 0 1pt 0}{
\begin{tikzpicture}[baseline=0.25mm]
    \fill[green] (0, 0) rectangle (0.25, 0.25);
    \draw[gray] (0,0) rectangle (0.25, 0.25);
\end{tikzpicture}
}
}

\DeclareRobustCommand{\lavaCell}{
\trimbox{1pt 0 1pt 0}{
\begin{tikzpicture}[baseline=0.25mm]
    \fill[orange] (0, 0) rectangle (0.25, 0.25);
    \draw[gray] (0,0) rectangle (0.25, 0.25);
\end{tikzpicture}
}
}

\DeclareRobustCommand{\checkpointEx}[2]{
\trimbox{1pt 0 1pt 0}{
\begin{tikzpicture}[scale=0.5, baseline=-5.5mm]
    \def\x{0}
    \def\y{-0.9}
    \def\radius{0.34}
    \fill[color=#1] (\x, \y) circle (\radius);
    \node[text=black, font=\scriptsize] at (\x, \y) {$#2$};
    \draw[gray] (\x, \y) circle (\radius);
\end{tikzpicture}
}
}

\DeclareRobustCommand{\agent}{
\def\agentpos{{0,0}}
\trimbox{0pt 0 2.5pt 0}{
\begin{tikzpicture}[scale=0.5, baseline=1.5mm]
\draw[fill=red] (0.9, 0.5) -- (0.1, 0.2) -- (0.1, 0.8) -- cycle;
\end{tikzpicture}
}
}

\newcommand{\runningExampleFig}{
\begin{figure*}
    \captionsetup[subfigure]{justification=centering}
    \captionsetup{justification=centering}
    \centering
    \begin{subfigure}[b]{0.33\textwidth}
        \centering
        \adjustbox{width=\textwidth}{
            \minigridTwoObj
        }
        \caption{Environment instance.}
        \Description[GridWorld environment Example]{Visualisation of the grid world environment where the agent must navigate and observe uniquely identified balls of different colors. In this example, the environment has two balls of each color: yellow, red, blue, purple, grey, green. The goal cell is a green square.}
        \label{fig:minigrid_ex}
    \end{subfigure}
    \hfill
    \begin{subfigure}[b]{0.65\textwidth}
        \raisebox{20pt}{
            \begin{subfigure}{0.98\textwidth}
                \centering
                \rmRunningExamplePropHor
                \caption{Traditional RM encoding of the example task.}
                \Description[Propositional RM Encoding example]{Example of a propositional RM Encoding of a task where the agent must visit all the yellow balls before visiting any of the blue balls to finally reach the goal cell.}
                \label{fig:prop_rm_task}
            \end{subfigure}
        }
        \vfill
        \begin{subfigure}{0.98\textwidth}
            \centering
            \rmRunningExampleFOLHor
            \caption{\FORM{} encoding of the example task.}
            \Description[\FORM{} Encoding example]{Example of a \FORM{} Encoding of a task where the agent must visit all the yellow balls before visiting any of the blue balls to finally reach the goal cell.}
            \label{fig:fol_rm_task}
        \end{subfigure}
    \end{subfigure}
    \caption{An instance of the environment (a), and RMs for the task \emph{``visit all $\checkpointEx{yellow}{}$ followed by any $\checkpointEx{blue}{}$  before reaching $\goalCell$''} (b, c). See Example~\ref{ex:running_example} for details.}
\end{figure*}
}
\newcommand\width{0.85\linewidth}

\newcommand{\formLearningResultsFig}{
\begin{figure*}
    \captionsetup[subfigure]{justification=centering}
    \captionsetup{justification=centering}
    \centering
    \begin{subfigure}[b]{\textwidth}
        \centering
        \captionsetup{justification=centering}
        \includegraphics[width=0.66\linewidth]{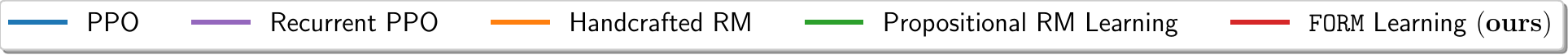}
        \Description[Legend]{Legend: PPO in blue, Handcrafted RM in orange, Propositional RM in green, \FORM{} (ours) in red}
    \end{subfigure}
    \hfill
    \begin{subfigure}[b]{0.33\textwidth}
        \centering
        \captionsetup{justification=centering}
        \includegraphics[width=\width]{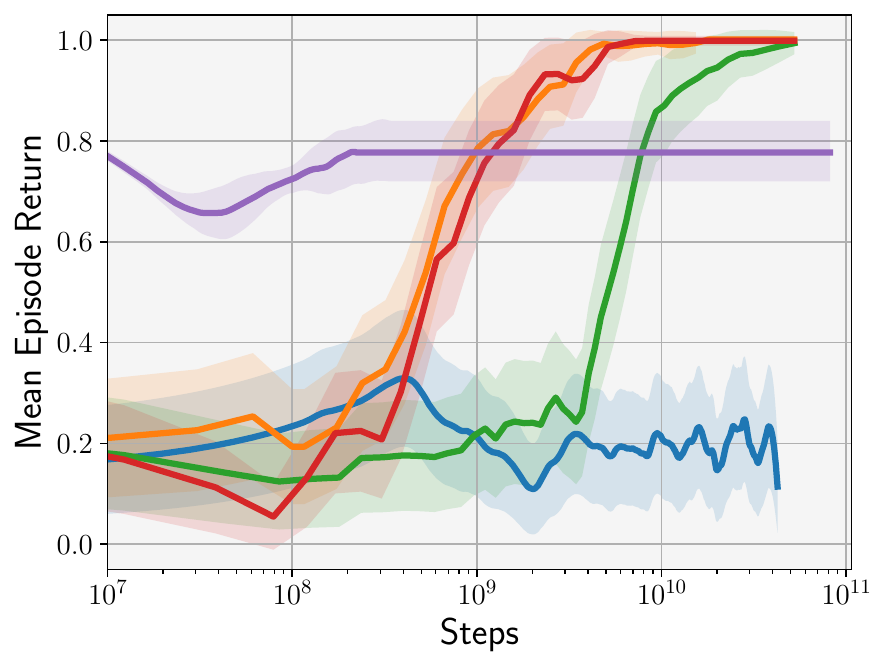}
        \caption{\textsc{AllYellow}}
        \Description[Mean Episode Return on the All Yellow task]{Our approach (where the \FORM{} is learnt) converges almost as quickly as when the RM is handcrafted and provided to the agent. The propositional RM learning method manages to converge after more iterations. PPO fails to learn how to solve the task.}
        \label{fig:all_yellow_results}
    \end{subfigure}
    \hfill
    \begin{subfigure}[b]{0.33\textwidth}
        \centering
        \captionsetup{justification=centering}
        \includegraphics[width=\width]{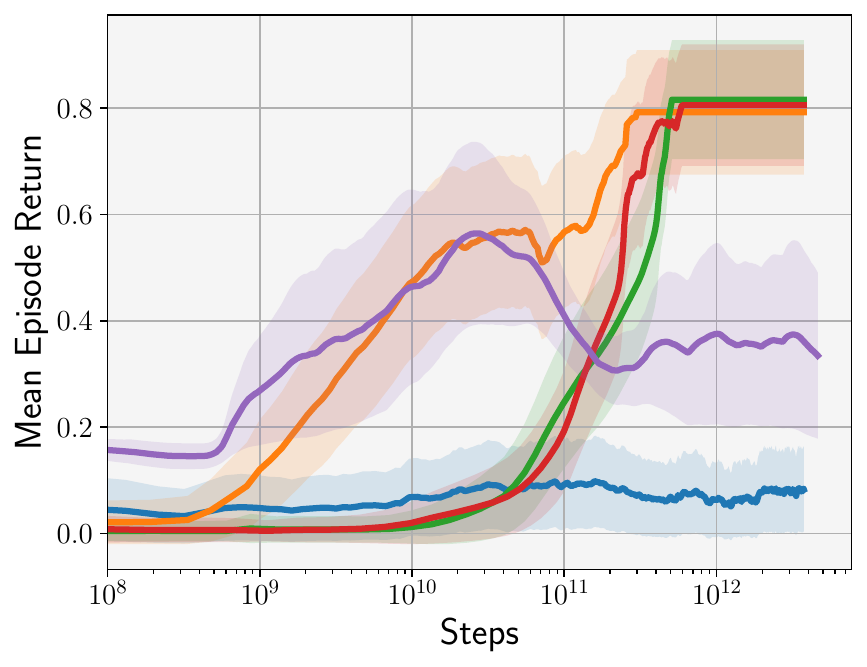}
        \caption{\textsc{GreenButOne-NoLava}}
        \Description[Mean Episode Return on the Green but One - No Lava]{The baseline with the handcrafter RM converges the fastest. Interestingly, the Propositional RM and \FORM{} converge almost as quickly. PPO fails to learn how to solve the task.}
        \label{fig:green_no_lava_results}
    \end{subfigure}
    \hfill
    \begin{subfigure}[b]{0.33\textwidth}
        \centering
        \captionsetup{justification=centering}
        \includegraphics[width=\width]{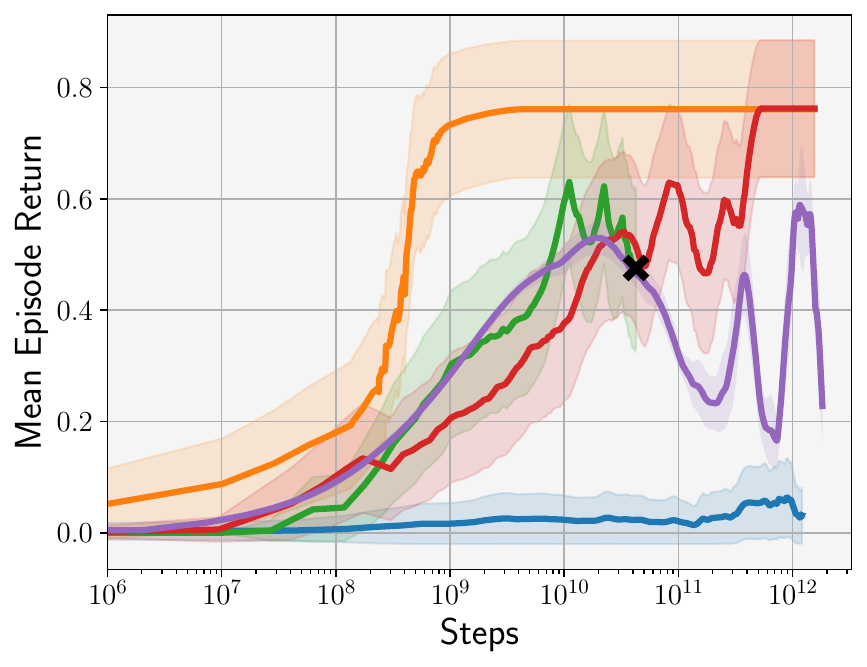}
        \caption{\textsc{Blue-AllYellow-7}}
        \Description[Mean Episode Return on the All Yellow task]{Our approach (where the \FORM{} is learnt) converges almost as quickly as when the RM is handcrafted and provided to the agent. The propositional RM learning method manages to converge after more iterations. PPO fails to learn how to solve the task.}
        \label{fig:blue_all_yellow_7_results}
    \end{subfigure}
    \caption{Average undiscounted return for the three tasks.}
\end{figure*}
}

\begin{document}


\pagestyle{fancy}
\fancyhead{}
 

\maketitle

\section{Introduction}



In recent years, reinforcement learning \cite[RL;][]{SuttonB18} has emerged as a powerful technique to train autonomous agents able to reach superhuman performance across a wide range of applications, including games \cite{mnih2015human, silver2016mastering,vinyals2019grandmaster,wurman2022outracing}, autonomous driving \cite{kendall2019learning,sallab2017deep}, finance \cite{spooner2018marketmaking,ardon2021towards}, and science \cite{senior2020improved,stokes2020deep}. Despite these successes, RL struggles to handle tasks that require long-term planning and abstraction, particularly when rewards are non-Markovian, i.e.~they depend on histories of states and actions.
Reward machines \cite[RMs;][]{icarte2018using} offer a promising solution by leveraging finite-state machines to encode the temporal structure of tasks and allow agents to handle non-Markovian rewards. However, traditional RMs rely on propositional logic to label transitions, which limits their expressivity and scalability. Abstracting over object properties (e.g., \emph{colour} or \emph{type}) is unfeasible with propositional logic: it requires all combinations to be encoded in the RM, leading to a combinatorial explosion in the number of states and edges. As a result, RMs are difficult to learn and transfer across different tasks, especially as the complexity of the tasks increases.

In this paper, we make the following contributions. \textbf{(i)} We introduce First-Order Reward Machines (\FORM{s}), a novel formulation of RMs that uses first-order logic to label transitions, enhancing expressivity and transferability. 
%
%
\textbf{(ii)} We propose a new RM learning method which is, to the best of our knowledge, the first to tackle the problem of \emph{learning} RMs with first-order logic.
\textbf{(iii)} We formalise the exploitation of an RM as a multi-agent problem, where multiple agents collaboratively learn the policies of the RM.
Finally, \textbf{(iv)} we demonstrate empirically the benefit of our approach over traditional RM learning methods, in both learning and transferability.

The paper is organised as follows. \secref{sec:background} introduces the background of our work. \secref{sec:methodology} formalises \FORM{}s, and describes our methods for learning and exploiting them. We evaluate our methods' performance and the reusability of \FORM{}s in \secref{sec:experiments}. \secref{sec:rw} discusses related work, and \secref{sec:concl} concludes the paper.

\runningExampleFig

\section{Background}\label{sec:background}
In this section, we cover three basic topics needed to understand our approach. We discuss reinforcement learning and a state-of-the-art algorithm which we use to learn policies. We present the notion of a reward machine which we build upon and generalise it to a First-Order Reward Machine (\FORM{}). Finally, we discuss the Learning from Answer Sets framework, which we use to learn \FORM{}s.

\subsection{Reinforcement Learning}
Reinforcement learning \cite[RL;][]{SuttonB18} is a machine learning paradigm where an agent learns to make sequential decisions by interacting with an environment, often modeled as a Markov decision process~(MDP). The agent's objective is to learn a policy $\pi$, a mapping from states to actions, that maximizes the expected cumulative discounted reward (called return) $R = \sum_{t=0}^{T} \gamma^t r_t$, where $r_t\in\mathbb{R}$ is the reward received at time-step $t$, $\gamma \in [0,1)$ is the discount factor, and $T$ is the time horizon. 
The agent observes the current state $s_t$ of the environment, takes an action $a_t \sim \pi(\cdot|s_t)$ according to its current policy $\pi$, transitions to a new state $s_{t+1} \sim p(\cdot|s_t, a_t)$ according to the environment transition function $p$, and receives a reward $r_t = r(s_t, a_t, s_{t+1})$ from the reward function $r$.

Proximal policy optimization~\cite[PPO;][]{schulman2017ppo} is a state-of-the-art RL algorithm that, based on the policy gradient theorem \cite{SuttonMSM99}, directly learns a policy $\pi(a_t|s_t)$ as an action distribution conditioned on a state $s_t$. PPO optimizes a surrogate objective function that balances exploration and exploitation; specifically, PPO uses a clipped probability ratio to constrain the policy update, ensuring that the new policy does not deviate too much from the old policy, preventing large and destabilizing changes. PPO has the advantage of being robust and effective on a wide range of tasks while requiring minimal hyperparameter tuning \cite{yu2022surprising}.

\subsection{Learning from Answer Sets}
\label{sec:background_learning_answer_sets}

Answer set programming \cite[ASP;][]{GelfondL88} is a declarative programming language for knowledge representation. A problem is expressed in ASP using logic rules, and the models (known as \emph{answer sets}) of its representation are its solutions. Within the context of this paper, an ASP \emph{program} $P$ is a set of \emph{normal} rules.  Given any atoms $\texttt{h}$, $\texttt{b}_\texttt{1},\ldots,\texttt{b}_\texttt{n}$, $\texttt{c}_\texttt{1},\ldots,\texttt{c}_\texttt{m}$, a normal rule is of the form $\texttt{h }\codeif \texttt{b}_\texttt{1},\ldots,\texttt{b}_\texttt{n},\texttt{not }\texttt{c}_\texttt{1},\ldots,\allowbreak\texttt{not }\texttt{c}_\texttt{m}$ where $\texttt{h}$ is the \emph{head},  $\texttt{b}_\texttt{1},\ldots,\texttt{b}_\texttt{n},\texttt{not }\texttt{c}_\texttt{1},\ldots,\texttt{not }\texttt{c}_\texttt{m}$ is (collectively) the \emph{body} of the rule, and ``$\texttt{not}$'' represents negation as failure.
An atom is \emph{ground} if it is variable free. 
%
%
%
%
Informally, given a set of ground atoms (or \emph{interpretation}) $I$, a ground normal rule is satisfied if the body is not satisfied by $I$ or the head is satisfied by $I$. The reader is referred to \cite{gelfond2014knowledge} for further details on the semantics of ASP programs.
%
%
%
\texttt{ILASP} \cite{ILASP_system} is a state-of-the-art inductive logic programming system for learning ASP programs from partial answer sets. 
%
%
A program $P$ is said to accept an example $e$ if and only if there exists an answer set $A$ of $P \cup e$.
An \texttt{ILASP} task~\cite{LawRB16} is a tuple $T=\langle B, S_M, E\rangle$, where $B$ is the ASP background knowledge, $S_M$ is the set of rules allowed in the hypotheses (called \emph{hypothesis space}), and $E$ is a set of 
%
examples
. A \emph{hypothesis} $H\subseteq S_M$ is a \emph{solution} of $T$ if and only if $B\cup H$ accepts all examples $e\in E$. 
%

\subsection{Reward Machines}
\label{sec:rms}
Reward machines \cite[RMs;][]{icarte2018using,Icarte_Klassen_Valenzano_McIlraith_2022} are finite-state machine representations of reward functions. RMs encode the temporal structure of a task's rewards, enabling expressive and interpretable specifications; besides, they enable handling non-Markovian reward tasks by using the state of the RM as an external memory.

Reward machines are defined in terms of high-level events (or \emph{observables}) $\observablesSet$, which aim to abstract the environment's state space $\envStatesSet$. More specifically, the transitions of an RM are labelled with sets of observables $\observation$ (or \emph{observations}). We denote by $\observationsSet$ the set of all possible observations, i.e.~the powerset of $\observablesSet$, such that $|\observationsSet| \ll |\envStatesSet|$. The abstraction from environment states and actions to observations is performed by a \emph{labelling function} $\labellingFunction: \envStatesSet \times \envActionsSet \times \envStatesSet \rightarrow \observationsSet$.

\begin{definition} [Reward Machine] \label{def:rm} 
An RM is a tuple $\RM = \langle \statesSet,\allowbreak \observablesSet,\allowbreak \uI,\allowbreak \uacc,\allowbreak \urej,\allowbreak \deltaF,\allowbreak \deltaR \rangle$, where $\statesSet$ is a finite set of states; $\observablesSet$ is a set of observables whose powerset is denoted as $\observationsSet$; $\uI, \uacc, \urej \in \statesSet$ are the initial, accepting and rejecting state, respectively; $\deltaF: \statesSet \times \observationsSet \rightarrow \statesSet$ is a deterministic state-transition function that maps a RM state and an observation to the next RM state; and $\deltaR: \statesSet \times \statesSet \rightarrow \mathbb{R}$ is a deterministic reward-transition function that returns the reward given a pair of RM states determining a transition.
\end{definition}

We now introduce a running example used throughout the paper.

\begin{example}\label{ex:running_example}
    Let us consider an agent ($\agent$) navigating the grid environment in \figref{fig:minigrid_ex}. The grid consists of several checkpoints (e.g.~$\checkpointEx{blue}{4}$), each characterised by a unique identifier and a colour, and a goal location $\goalCell$ placed in the bottom right corner. These checkpoints and the goal constitute the set of observables $\observablesSet$. At each step, the agent observes the state of the environment. The labelling function $\labellingFunction$ returns the set of observables in the cell where the agent has stepped on, e.g.~$\{\checkpointEx{blue}{4}\}$ if the agent moves forward in \figref{fig:minigrid_ex}.

    Several non-Markovian tasks can be formulated in this environment, each consisting of observing a sequence of observables, e.g.~``visit all yellow checkpoints $\checkpointEx{yellow}{}$ followed by any blue checkpoint $\checkpointEx{blue}{}$ before reaching $\goalCell$''. 
    \figref{fig:prop_rm_task} shows the \emph{propositional} RM for this task. To ``visit all yellow checkpoints'', the agent may start with $\checkpointEx{yellow}{0}$ followed by $\checkpointEx{yellow}{1}$ or vice-versa. To ``visit any blue checkpoints'', the agent can visit either $\checkpointEx{blue}{4}$ or $\checkpointEx{blue}{5}$. After completing these  sub-tasks, in the specified order, the agent can reach the goal $\goalCell$.
\end{example}

We \emph{assume} the tasks are such that the reward is $1$ if completed and $0$ otherwise. Accordingly, in the case of RMs, we assume the reward is $1$ only on transitions to the accepting state $\uacc$. We thus omit the rewards from the RM figures for clarity.

\subsubsection*{Learning of Reward Machines.}
\label{sec:background_learn_rms}
When the RM is unknown `a priori', it can be learnt from sequences of observations (or \emph{traces}) seen by the agent.
\citet{Furelos-Blanco_Law_Jonsson_Broda_Russo_2021} present an approach that learns RMs from traces using the \texttt{ILASP} inductive logic programming system introduced in \secref{sec:background_learning_answer_sets}. The RM is expressed in ASP using the following rules:
\begin{itemize}
    \item Facts $\texttt{ed}(u,u',i)$, which encode the presence of an edge from $u$ to $u'$ with identifier $i$; and
    \item Rules with head $\bar{\transitionFormulaFunction}(u,u',i,\texttt{T)}$, which encode the negation of the formulae to be satisfied for the RM transition $i$ from $u$ to $u'$.
\end{itemize}

\begin{example}\label{ex:asp_rule}
    Given the RM in \figref{fig:prop_rm_task}, the ASP rules that encode the transition $\transition{u_3}{u_4}$ are:
    \begin{gather*}
\small
\left\lbrace\begin{array}{@{}l}
\texttt{ed}(u_3,u_4,0).
\\
\texttt{ed}(u_3,u_4,1).
\\
\bar{\transitionFormulaFunction}(u_3,u_4,0,\texttt{T)} \codeif \texttt{not~obs}(\checkpointEx{blue}{4}, \texttt{T), step(T).} 
\\
\bar{\transitionFormulaFunction}(u_3,u_4,1,\texttt{T)} \codeif \texttt{not~obs}(\checkpointEx{blue}{5}, \texttt{T), step(T).} 
\\
\end{array}\right\rbrace,
\end{gather*}

\noindent
    where atoms $\texttt{obs}(o,\texttt{T})$ and $\texttt{step(T)}$  indicate respectively that observable $o$ is observed at time $\texttt{T}$, and $\texttt{T}$ is a time-step. Note that (i)~the disjunction is encoded with two edges whose indices are $0$ and $1$, and (ii)~the transition rules encode $\bar{\transitionFormulaFunction}$ (i.e., the negation of $\transitionFormulaFunction$), explaining the use of $\texttt{not}$ \cite{Furelos-Blanco_Law_Jonsson_Broda_Russo_2021}.

\end{example}

The hypothesis space for the \texttt{ILASP} task learning the ASP rules that encode the reward machine, like the ones from \exref{ex:asp_rule}, is defined as:
\begin{gather}\label{eq:hypothesis_space_prop_rm}
\small
S_M=\left\lbrace\begin{array}{@{}l}
\texttt{ed}(u,u',i).
\\
\bar{\transitionFormulaFunction}(u,u',i,\texttt{T)} \codeif \texttt{obs}(o, \texttt{T), step(T).} 
\\
\bar{\transitionFormulaFunction}(u,u',i,\texttt{T)} \codeif \texttt{not~obs}(o, \texttt{T), step(T).}
\end{array}\right\rbrace,
\end{gather}
where $u \in \statesSet \setminus \left\lbrace \uacc,\urej \right\rbrace$, $u' \in \statesSet \setminus \left\lbrace u \right\rbrace$, $i \in \left[1, \maxDisjuncts \right]$, $o \in \observablesSet$, and $\maxDisjuncts$ is a hyperparameter denoting the maximum number of disjuncts (i.e.,~edges) between each pair of states. Traces, which form the examples of an \texttt{ILASP} task, are encoded as a set of $\texttt{obs(}o, \texttt{T)}$ facts; e.g., $\{\texttt{obs(}\goalCell\texttt{,0)$.  $obs(}\checkpointEx{yellow}{1}\texttt{,1).step(0).step(1)}\}$. To be considered as a solution, a hypothesis (i.e., the ASP encoding of an RM) must accept all traces. 

As it can be observed in \eqref{eq:hypothesis_space_prop_rm}, the number of $\texttt{ed}$ facts in the hypothesis space scales quadratically with the number of states in the RM ($|\statesSet|$) and linearly with the maximum of disjuncts allowed in the transition rules ($\maxDisjuncts$): $(|\statesSet | - 2) \cdot (|\statesSet | - 1) \cdot \maxDisjuncts$. Additionally, the number of $\bar{\transitionFormulaFunction}$ rules depends on the number of observables ($|\observablesSet|$): $(|\statesSet | - 2) \cdot (|\statesSet | - 1) \cdot \maxDisjuncts \cdot 2 |\observablesSet|$. As the number of states, disjuncts, or observables increase, learning the RM quickly becomes unfeasible. This underscores the need for new solutions to ensure that RM learning can scale effectively.

\section{Methodology}\label{sec:methodology}
The limited expressivity of propositional logic formulae presents significant challenges to the broader adoption of RM-based approaches for RL. For instance, propositional logic cannot express tasks in terms of the properties of a specific or a group of entities, such as the colour of checkpoints in Example~\ref{ex:running_example}. This limitation substantially increases the number of sub-task combinations to be captured (if expressed at propositional level), resulting in larger RMs that are more difficult to learn and exploit. In addition, the restriction to a fixed set of observables hinders the re-use of RMs across different scenarios (e.g.,~on environments with different numbers of objects of the same colour). To address these limitations, we propose to label the edges of RMs using \emph{first-order} formulae. The high expressivity of first-order logic allows for more compact and general RMs, making them easier to learn and transfer across different scenarios. We first illustrate the intuition behind our approach by building upon the RM from \exref{ex:running_example}.

\begin{example}\label{ex:running_example_2}
    \figref{fig:fol_rm_task} displays the First-Order RM (\FORM) for the task ``visit all yellow checkpoints $\checkpointEx{yellow}{}$ followed by any blue checkpoint $\checkpointEx{blue}{}$ before reaching $\goalCell$''. Despite encoding the same task, the \FORM{} is more compact than its propositional counterpart~(\figref{fig:prop_rm_task}).

    Existentially quantified formulae reduce the number of edges required to encode the disjunctions, impacting the parameter $\maxDisjuncts$ in \eqref{eq:hypothesis_space_prop_rm} (\figref{fig:existential_motivation}).
    Universally quantified formulae reduce the number of states and, consequently, the number of edges required to encode a task (\figref{fig:universal_motivation}).\footnote{An example with three yellow checkpoints is shown in \appendixref{apdx:additional_examples}.}
    \motivationFig{h}
\end{example}

We now formalise the first-order language over which the transitions of \FORM{}s are defined (Section~\ref{sec:language_interpretation}) and provide a formal definition of a first-order RM (Section~\ref{sec:forms}). We then introduce our method for learning \FORM{}s (Section~\ref{sec:form_learning}) and exploiting them (Section~\ref{sec:form_policy_learning}).

\subsection{Language}\label{sec:language_interpretation}
In this section, we introduce the first-order language $\firstOrderLanguage$ used to label transitions in \FORM{}s. The \emph{signature} of $\firstOrderLanguage$, denoted as $\Sigma = \langle \domain, \predicateSet \rangle$, is composed of a set $\domain$ of \emph{constants} and a set $\predicateSet$ of $n$-ary predicates. Predicates with $0$-arity are referred to as \emph{propositions}. A variable $X$ of a predicate $\pred{P} \in \predicateSet$ is \emph{bound} if it is in the scope of a quantifier $\quantifier \in \quantifierSet$. A \emph{ground atom} is a predicate whose arguments are constants.
A \textit{quantified atom} is a predicate whose variables are bound, e.g.~$\exists X. \predvar[P][]{X}$. A \emph{ground instance}, $\psi_g$, of a quantified atom $\psi$ is a ground atom generated by replacing all variables in $\psi$ with constants. Propositions, ground atoms and quantified atoms are \emph{atomic formulae} of $\firstOrderLanguage$. The first-order language $\firstOrderLanguage$ of a \FORM{} is defined as follows.

\begin{definition}
\label{def:grammar}
Given a signature $\Sigma = \langle \domain, \predicateSet \rangle$, the language $\firstOrderLanguage$ is the set of \emph{formulae} defined inductively as follows:
\begin{enumerate}
    \item An atomic formula is a formula.
    \item If $\psi$ is a formula, then so is $\neg\psi$.
    \item If $\psi_1$ and $\psi_2$ are formulae, then $\psi_1 \wedge \psi_2$, and $\psi_1 \vee \psi_2$ are formulae.
\end{enumerate}
We refer to formulae in a first-order language $\firstOrderLanguage$ as $\firstOrderLanguage$-formulae.
\end{definition}

The set of all propositions and ground atoms in $\firstOrderLanguage$ is called the Herbrand Base of $\firstOrderLanguage$, denoted $\herbrandBase$.

\begin{example}
   Consider the environment given in \figref{fig:minigrid_ex}. The language $\firstOrderLanguage$ for this environment has the signature $\Sigma = \langle \domain, \predicateSet \rangle$ given by the following sets:
    \begin{align*}
        \domain =& \{ o_0, \ldots, o_{12} \}, \\
        \predicateSet =& \{ \checkpointEx{yellow}{}, \checkpointEx{blue}{}, \checkpointEx{red}{}, \checkpointEx{gray}{}, \checkpointEx{purple}{}, \checkpointEx{green}{}, \goalCell\},
    \end{align*}
    where the predicates are pictograms for unary predicates $\pred{yellow}$, $\pred{blue}$, $\pred{red}$, $\pred{gray}$, $\pred{purple}$, $\pred{green}$, and the proposition $\pred{goal}$, respectively. We depict a ground atom of $\firstOrderLanguage$ as a predicate pictogram with the subscript of its constant argument, e.g. $\pred{blue}(o_4)$ is depicted as $\checkpointEx{blue}{4}$. Analogously, we depict quantified atoms with the quantifier and the predicate pictogram with the bound variable, e.g. $\forall X.\pred{yellow}(X)$ is depicted as $\forall X;. \checkpointEx{yellow}{X}$. The proposition $\goalCell$ and the negated formula $\neg\checkpointEx{blue}{4}$ are examples of $\firstOrderLanguage$-formulae.
\end{example}


\subsection{First-Order Reward Machines (\FORM{}s)}\label{sec:forms}
We formalise now the notion of a first-order RM (\FORM{}), where transitions are labelled with formulae of a first-order language $\firstOrderLanguage$. \FORM{}s enhance the expressivity of traditional RMs. The core entities that characterise a \FORM{} are its observables, labelling function, and state-transition function. In what follows, we describe how these entities are defined and how the satisfiability of $\firstOrderLanguage$-formulae labelling the transitions of a \FORM{} is checked.

\begin{definition} [First-Order RM] \label{def:folrm} 
A First-Order RM over a first-order language $\firstOrderLanguage$ is a tuple $\FORM_{\firstOrderLanguage} = \langle \statesSet, \herbrandBase, \uI, \uacc, \urej, \deltaF, \deltaR \rangle$ where $\statesSet, \uI, \uacc, \urej$ and $\deltaR$ are as defined in \defref{def:rm}, and 
\begin{itemize}
   \item $\herbrandBase$ is the set of observables, and
   \item $\deltaF:\statesSet\times \observationsSet^* \to \statesSet \times \observationsSet^*$ is the state-transition function taking a state and a history of observations to a new state and an updated history.
\end{itemize}

\end{definition}

\subsubsection*{Observables and Labelling Function}
Differently from traditional RMs, where \emph{observables} are given as a set of propositions $\observablesSet$, the observables in a $\FORM_{\firstOrderLanguage}$ are defined as the Herbrand base of the first-order language $\firstOrderLanguage$. The \emph{labelling function} $\labellingFunction$ thus returns at each time-step, a subset of $\herbrandBase$ as an observation.

\subsubsection*{State-Transition Function}
The \emph{state-transition function} $\deltaF$ generalises that of a traditional RM in two ways. Firstly, it supports the use of first-order formulae returned by a logical transition function $\transitionFormulaFunction:\statesSet \times \statesSet \to \firstOrderLanguage$. Secondly, it takes a \emph{history of observations} instead of a single observation.
This is important to semantically evaluate universally quantified atoms.
To relax the requirement of having to see \emph{all} the ground instances of the universally quantified atoms in the \emph{same} observation, we use the following interpretation. Given a history, a universally quantified atom is \true when all its ground instances are contained in the history.
We refer to this history as a \emph{buffer} $\buffer{} = [ \observation_{t-k}, \ldots, \observation_t ]$ which gathers observations from time-step $t-k$ to $t$, where $k \geq 0$ is the length of the buffer.

\algoref{alg:deltaF} shows the pseudo-code that defines $\deltaF$. The buffer $\buffer{}$ contains all the observations perceived since the current RM state was reached. The buffer is emptied when a transition to a new state is taken (\algolineref{algl:buffer_empty}), or updated with the current observation otherwise (\algolineref{algl:buffer_aggregation}). The $\firstOrderLanguage$-formulae are given by the logical transition function $\transitionFormulaFunction$ and are evaluated against the buffer (\algolineref{algl:buffer_evaluation}).

\begin{algorithm}[h]
    \caption{State-Transition Function $\deltaF$}
    \label{alg:deltaF}
    \begin{flushleft}
    \hspace*{\algorithmicindent}\textbf{Input:} The current \FORM{} state $u$, the latest observation $\observation_t$, the buffer $\buffer{}$. \\
    \hspace*{\algorithmicindent}\textbf{Output:} The next \FORM{} state $u'$, the updated buffer .
    \end{flushleft}
    \begin{algorithmic}[1]
        \STATE $\buffer{} \gets \buffer{} \oplus \observation_t$ \label{algl:buffer_aggregation}
        \FORALL{$u' \in \statesSet$}
            \IF{$\buffer{} \models \transitionFormulaFunction(u, u')$} \label{algl:buffer_evaluation}
                \STATE $\buffer{}\gets []$ \label{algl:buffer_empty}
                \STATE \textbf{return} $u'$, $\buffer{}$
            \ENDIF
        \ENDFOR
        \STATE $u' \gets u$
        \STATE \textbf{return} $u'$, $\buffer{}$
    \end{algorithmic}
\end{algorithm}

In the following section, we define the notion of  satisfiability of an $\firstOrderLanguage$-formula given a buffer $\buffer{}$.

\subsubsection*{Satisfiability of $\firstOrderLanguage$-formulae}
Given a buffer $\buffer{}=[\observation_{t-k},\ldots,\observation_t]$, observables that are included in $\buffer{}$ are assumed to be \true. Any other observable from $\herbrandBase$ not included in $\buffer{}$ is assumed to be \false. We can now define the satisfiability of a $\firstOrderLanguage$-formula given a buffer $\buffer{}=[\observation_{t-k},\ldots,\observation_t]$.

\begin{definition}
Let $\buffer{}=[\observation_{t-k},\ldots,\observation_t]$ be a buffer for some $k\geq 0$. Let $\psi$ be a $\firstOrderLanguage$-formula. $\buffer{}$ satisfies $\psi$, written $\buffer{} \models \psi$, is defined as follows:
\begin{itemize}
\item $\buffer{} \models \psi$ if $\psi\in \observation_t$, where $\psi$ is a proposition or a ground atom.
\item $\buffer{} \models \psi$ if $\psi_g\in \observation_t$ for \emph{some} ground instance $\psi_g$ of $\psi$, where $\psi$ is an \textbf{existentially} quantified atom. 
\item $\buffer{} \models \psi$ if $\psi_g\in \bigcup_{0 \leq i \leq k} \observation_{t-i}$ for \emph{all} ground instance $\psi_g$ of $\psi$, where $\psi$ is a \textbf{universally} quantified atom. 
\item $\buffer{} \models \neg \psi$ if $\buffer{} \not\models\psi$
\item $\buffer{} \models \psi_1\wedge\psi_2$ if $\buffer{}\models\psi_1$ and $\buffer{}\models\psi_2$.

\end{itemize}
\end{definition}

Note that a universally quantified formula is satisfied over a buffer when all ground instances of the formula, in the language $\firstOrderLanguage$ have been observed, i.e. are included in the buffer. The number of occurrences of the same observation throughout the buffer is not relevant. All other atomic formulae of $\firstOrderLanguage$ are satisfied over a buffer, if the most recent observation satisfies them. 

\begin{example}
Given the RM in \figref{fig:fol_rm_task}, we exemplify the satisfiability of different $\firstOrderLanguage$-formulae $\psi$ for different buffer examples:
\begin{description}
    \item[Proposition] Let $\psi = \goalCell$ and $\bufferSymbol=[\{\goalCell\} \big]$; then, $\bufferSymbol\models\psi$.
    \item[Existentially Quantified Atom] Let $\psi = \exists X\checkpointEx{blue}{X}$ and $\bufferSymbol=\allowbreak\big[ \{\checkpointEx{yellow}{0}\},\allowbreak \{\checkpointEx{blue}{4}\} \big]$; then, $\bufferSymbol\models\psi$.
    \item[Universally Quantified Atom] Let $\psi = \forall X. \checkpointEx{yellow}{X}$, $\bufferSymbol_1 = \big[ \{\goalCell\},\allowbreak \{\checkpointEx{yellow}{0}\},\allowbreak \{\checkpointEx{blue}{4}\},\allowbreak \{\checkpointEx{yellow}{1}\} \big]$, and $\bufferSymbol_2 = \big[ \{\goalCell\}, \{\checkpointEx{yellow}{0}\}, \{\checkpointEx{blue}{4}\} \big]$; then, $\bufferSymbol_1\models\psi$ and $\bufferSymbol_2\not\models\psi$.

    \item[Conjunctions] Let $\psi = \forall X.\checkpointEx{yellow}{X} \wedge \goalCell$, $\bufferSymbol_1 = \big[ \{\goalCell\},\allowbreak  \{\checkpointEx{yellow}{0}\},\allowbreak \{\checkpointEx{blue}{4}\},\allowbreak \emptyset,\allowbreak \{\checkpointEx{yellow}{1}\} \big]$, and $\bufferSymbol_2=\big[ \{\goalCell\},\allowbreak \{\checkpointEx{yellow}{0}\},\allowbreak \{\checkpointEx{blue}{4}\},\allowbreak \emptyset,\allowbreak \{\checkpointEx{yellow}{1}\},\allowbreak \emptyset,\allowbreak \{\goalCell\} \big]$; then, $\bufferSymbol_1 \not\models \psi$  since $\bufferSymbol_1 \models \forall X. \; \checkpointEx{yellow}{X}$ but $\bufferSymbol_1 \not\models \goalCell$, and $\bufferSymbol_2 \models \psi$ since $\bufferSymbol_2 \models \forall X. \; \checkpointEx{yellow}{X}$ and $\bufferSymbol_2 \models \goalCell$.
\end{description}

\end{example}

\subsection{\FORM{} Learning}
\label{sec:form_learning}
We now focus on the problem of learning a \FORM{} from the observation traces collected by the agent. Although observations are sets of ground atoms and propositions, the goal is to learn an abstract representation of the task's structure using lifted first-order formulae. To achieve this, we leverage the support of ASP for variables and learn ASP rules for the $\firstOrderLanguage$-formulae constituting the \FORM{}.

The set of observables ($\herbrandBase$) and the set of predicates ($\predicateSet$) are \emph{not} known `a priori'; instead, they are automatically derived from the observation traces \update{and used to dynamically generate a new ILASP learning task.} We model existentially and universally quantified atoms as ASP rules and expand the hypothesis space from \eqref{eq:hypothesis_space_prop_rm} to allow the edges of the RM to be labelled with quantified atoms or their negations. Formally, 
\begin{gather*}\label{eq:hypothesis_space_form}
\small
S_M=\left\lbrace\begin{array}{@{}l}
\texttt{ed}(u,u',i).
\\
\bar{\transitionFormulaFunction}(u,u',i,\texttt{T)} \codeif \texttt{obs}(o, \texttt{T),step(T)}. 
\\
\bar{\transitionFormulaFunction}(u,u',i,\texttt{T)} \codeif \texttt{not~obs}(o, \texttt{T),step(T)}.
\\
\bar{\transitionFormulaFunction}(u,u',i,\texttt{T)} \codeif \texttt{e\_pred}(p, \texttt{T),step(T)}. 
\\
\bar{\transitionFormulaFunction}(u,u',i,\texttt{T)} \codeif \texttt{not~e\_pred}(p, \texttt{T),step(T)}.
\\
\bar{\transitionFormulaFunction}(u,u',i,\texttt{T)} \codeif \texttt{a\_pred}(p, \texttt{T),step(T)}. 
\\
\bar{\transitionFormulaFunction}(u,u',i,\texttt{T)} \codeif \texttt{not~a\_pred}(p, \texttt{T),step(T)}.
\end{array}\right\rbrace,
\end{gather*}
where $u \in \statesSet \setminus \left\lbrace \uacc,\urej \right\rbrace$, $u' \in \statesSet \setminus \left\lbrace u \right\rbrace$, $i \in \left[1, \kappa \right]$, $o \in \herbrandBase$, and $p \in \predicateSet$.

\lstref{lst:existential} presents the ASP encoding for an \emph{existentially quantified atom}.  The rule $\texttt{e\_pred(\textcolor{\predicateListingColor}{p},T)}$ is \true iff a ground atom of $\texttt{\textcolor{\predicateListingColor}{p}}$ is observed at time-step $\texttt{T}$. \lstref{lst:universal} shows the ASP encoding of a \emph{universally quantified atom}. The rule $\texttt{\textcolor{\predicateListingColor}{p}\_holds(O,T)}$ is \true whenever the observable $\texttt{O}$ is present in the \emph{buffer} at time-step $\texttt{T}$. We use this rule to encode the $\texttt{all\_\textcolor{\predicateListingColor}{p}\_hold(T)}$, evaluating to \true for all time-steps $\texttt{T}$ for which all the ground atoms of $\texttt{\textcolor{\predicateListingColor}{p}}$ holds. Finally, the rule $\texttt{a\_pred(\textcolor{\predicateListingColor}{p},T)}$, used in the hypothesis space, encodes a universally quantified atom and is \true for the first time-step $\texttt{T}$ when all the ground atoms of $\texttt{\textcolor{\predicateListingColor}{p}}$ have been observed. These rules make use of the atoms:
\begin{itemize}
    \item $\texttt{obs(O, T)}$: the observable $\texttt{O}$ is seen at time-step $\texttt{T}$;
    \item $\texttt{st(T, X)}$: the RM was in state $\texttt{X}$ at time-step $\texttt{T}$;
    \item $\texttt{step(T)}$: $\texttt{T}$ is a valid time-step (i.e. its value is between $0$ and the maximum trace length); and
    \item $\texttt{state(X)}$: $\texttt{X}$ is an RM state.
\end{itemize}

\begin{listing}[h]%
\caption{ASP encoding for an existentially quantified sentence $\exists X. {\color{\predicateListingColor}\texttt{p}}(X)$}%
\lstset{numberstyle=\tiny, escapeinside={(*@}{@*)}}
\label{lst:existential}%
\begin{lstlisting}[title={\textit{The rule $\texttt{e\_pred(}\textcolor{\predicateListingColor}{\texttt{p}}, \texttt{T)}$ is \true when an observable seen at time-step $\texttt{T}$ is a ground atom of $\textcolor{\predicateListingColor}{\texttt{p}}$.}}, captionpos=b]
e_pred((*@\textcolor{\predicateListingColor}{\texttt{p}}@*), T) :- obs(O, T), (*@\textcolor{\predicateListingColor}{p}@*)(O).
\end{lstlisting}
\end{listing}

\begin{listing}[h]%
\caption{ASP encoding for a universally quantified sentence  $\forall X. {\color{\predicateListingColor}\texttt{p}}(X)$}%
\lstset{numberstyle=\tiny, escapeinside={(*@}{@*)}, morecomment=[f][\color{\listingColor}][0]{\#}}
\label{lst:universal}%
\begin{lstlisting}[title={\textit{The rule $\texttt{a\_pred(}\textcolor{\predicateListingColor}{\texttt{p}}\texttt{, T)}$ is \true when all the ground atoms of $\textcolor{\predicateListingColor}{\texttt{p}}$ have all been observed while being in the same RM state.}}, captionpos=b]
# evaluate if the observation (*@\textcolor{\predicateListingColor}{p}@*)(O) was observed at (*@\textcolor{black}{T}@*) 
# while being in the current state of the RM.
(*@\textcolor{\predicateListingColor}{p}@*)_holds(O, T) :- 
    obs(O, T2), (*@\textcolor{\predicateListingColor}{p}@*)(O),
    T >= T2, st(T, X), st(T2, X), 
    step(T), step(T2), state(X).
# rules to evaluate whether all the ground atoms 
# of (*@\textcolor{\predicateListingColor}{p}@*) hold at (*@\textcolor{black}{T}@*).
not_all_(*@\textcolor{\predicateListingColor}{p}@*)_hold(T) :- 
    not (*@\textcolor{\predicateListingColor}{p}@*)_holds(O, T), (*@\textcolor{\predicateListingColor}{p}@*)(O), step(T).
all_(*@\textcolor{\predicateListingColor}{p}@*)_hold(T) :- 
    not not_all_(*@\textcolor{\predicateListingColor}{p}@*)_hold(T), step(T).
# rule used in the hypothesis space that will be true
# when all the ground atoms of (*@\textcolor{\predicateListingColor}{p}@*) have been seen for
# the first time in the current state of the RM
a_pred((*@\textcolor{\predicateListingColor}{p}@*), T) :- all_(*@\textcolor{\predicateListingColor}{p}@*)_hold(T), 
    not all_(*@\textcolor{\predicateListingColor}{p}@*)_hold(T-1), step(T).
\end{lstlisting}
\end{listing}

\formLearningResultsFig

\subsubsection*{Determinism}
The state-transition function $\deltaF$, like in traditional RMs, must be deterministic; that is, two or more transitions cannot be simultaneously satisfied from a given state. To guarantee this property, the $\firstOrderLanguage$-formulae on such transitions must be \emph{mutually exclusive}. Two transitions are mutually exclusive if an atomic formula appears positively on one transition and negatively on the other.

Mutual exclusivity is encoded through ASP rules and enforced at learning time so that the output \FORM{}s are guaranteed to be deterministic. Since quantified atoms can represent multiple associated ground instances, we also define rules that translate quantified atoms into their corresponding sets of ground instances. We refer the reader to \appendixref{apdx:determinism_encoding} for the encoding of mutual exclusivity and the mapping from quantified atoms into ground instances.

\subsection{Policy Learning}
\label{sec:form_policy_learning}

The vast majority of the algorithms used to learn the policy associated with the RM are based on CRM~\cite{Icarte_Klassen_Valenzano_McIlraith_2022}, an algorithm that learns a \emph{single} policy for the whole RM. This poses some concerns on the transferability of the learnt solution, and more specifically on the ability to reuse the policy associated with a given sub-task of the RM.
To address this issue, we propose a novel formulation as a \emph{multi-agent problem}, where a team of agents (one for each RM state) collaborate to complete the task. Our approach grounds the problem of RL with RM into the Markov game framework and, combined with \FORM{}, enables the transfer of the learnt policies across scenarios (e.g., with different objects) to speed up the learning.


We formalise the problem as a collaborative \emph{Markov game} of $N$ agents defined with the tuple $\mathcal{G} = \langle \mathcal{RM}, N, \bm{\envStatesSet}, \bm{s_I}, \bm{\envActionsSet}, p, r, \gamma \rangle$, where $\mathcal{RM}$ is a \FORM{}, $N = |\statesSet_{\mathcal{RM}}|$ is the number of agents equal to the number of states in the \FORM{}; $\bm{\envStatesSet} = \envStatesSet'_1 \times \cdots \times \envStatesSet'_{N}$ is the set of joint states; $\bm{s_I} \in \bm{\envStatesSet}$ is a joint initial state; $\bm{\envActionsSet} = \envActionsSet_1 \times \cdots \times \envActionsSet_N$ is the set of joint actions; $p: \bm{\envStatesSet} \times \bm{\envActionsSet} \to \bm{\envStatesSet}$ is a deterministic joint transition function; and $r:(\bm{\envStatesSet} \times \bm{\envActionsSet})^+ \times \bm{\envStatesSet} \to \mathbb{R}^{N}$ is the joint reward function. The objective of the game is to find a team policy $\bm{\pi} = \pi_1 \times \cdots \times \pi_N$ mapping joint states to joint actions such that the sum of the individual expected cumulative rewards is maximised.

In practice, each state of the \FORM{} is assigned an RL agent responsible for learning a policy to accomplish the associated sub-task. Since the \FORM{} state-transition function is deterministic, we can only be in a single state at any given time; therefore, only one RL agent acts at each time-step. The joint reward function is designed so that the reward received is uniformly shared among all the agents that contributed to reaching the current state (i.e., $\frac{r}{n'}$ where $r$ is the global reward obtained and $n'$ is the number of agents that have participated in getting $r$), implying that their assigned \FORM{} state was visited. Consequently, the agents work \emph{collaboratively} and \emph{in turns} to traverse the \FORM{} towards the accepting state.

\subsubsection*{Agent State Space}
Since all agents act in the same environment, their state space $\envStatesSet'_u$ is the environment's state space $\envStatesSet$. However, depending on the outgoing transitions from a given \FORM{} state $u \in \statesSet$, the state space $\envStatesSet'_u$ of the associated agent may be extended. As described in Section~\ref{sec:forms}, universally quantified formulae are evaluated over multiple time-steps using an observation buffer $B$, making the sub-task non-Markovian as the policy becomes dependent on a history (i.e., the buffer). 

We address this issue by extending the state space $\envStatesSet'_u$ of agent $u$ with indicators encoding whether the ground atoms of interest have been perceived by the agent, hence making the sub-task Markovian. Let $g_u$ denote the set of ground atoms of a universally quantified formula that must be satisfied to exit a \FORM{} state $u$. The extended state $s'_u$ is constructed by concatenating the state $s$ from the environment with the indicators computed using the buffer $\buffer{}$:
\begin{gather*}
    s'_u = s \oplus [ \mathds{1}_{\{p \in \buffer{}\}} \; ; \forall p \in g_u ].
\end{gather*}

\begin{example}
    Let us consider $\transitionFormulaFunction_{\transition{u}{u'}} = \forall X. \checkpointEx{blue}{X} \vee \checkpointEx{red}{2} \;$. The steps followed to extend the state space are:
    \begin{enumerate}
        \item Identify the universally quantified predicates: $\{\checkpointEx{blue}{}\}$.
        \item Get the ground atoms associated: $g_u = \big[\checkpointEx{blue}{4}, \checkpointEx{blue}{5}\big]$.
        \item Extend the state space: $\envStatesSet'_u = \envStatesSet \times \{0, 1\}^{|g_u|}$.
    \end{enumerate}
    At time-step $t=5$, assume $\buffer{} = \big[ \{\checkpointEx{yellow}{0}\}, \{\checkpointEx{blue}{4}\} \big]$; then, $s'_{u, 5} = s_{5} \oplus [1, 0]$.
\end{example}

The \FORM{} states without outgoing transitions labelled with universally quantified formulae keep the state $\envStatesSet$ of the environment.

\section{Experiments}\label{sec:experiments}
We demonstrate the benefits of the proposed \FORM{} learning and exploitation approach in the environment from \figref{fig:minigrid_ex}. The tasks posed in this environment are challenging to traditional RM learning methods, especially as the number of objects increases. We first show the benefits of \FORM{} learning with respect to standard RL algorithms and methods leveraging propositional RMs (with and without learning them). Second, we show that first-order logic eases the transfer of RMs and policies to more complex scenarios (e.g.,~with more objects). Additional experimental details are described in \appendixref{apdx:experiment_details}.\footnote{\update{Code available at \url{https://github.com/leoardon/form}.}}

\subsection{\FORM{} Learning}
We evaluate the performance of our approach in its ability to learn and exploit \FORM{}s. We consider three baselines: (i)~PPO \cite{schulman2017ppo}, to show how RMs help deal with non-Markovian tasks; (ii)~Recurrent PPO using a recurrent network to keep track of the history (iii)~Handcrafted \FORM{}, which assumes the \FORM{} is known `a priori'; and (iv)~a state-of-the-art propositional RM learning approach \cite{Furelos-Blanco_Law_Jonsson_Broda_Russo_2021}. Both (iii) and (iv) employ our multi-agent based exploitation method.

In the following paragraphs, we compare our approach against the baselines in three, increasingly complex, tasks. 

\subsubsection*{\textbf{Task 1 -- \textsc{AllYellow}}} The task consists in visiting all yellow checkpoints $\checkpointEx{yellow}{}$ before going to the $\goalCell$ location. \figref{fig:minigrid_ex} illustrates the environment with two yellow checkpoints. This task is chosen to be relatively simple to ensure that traditional RM learning methods work so that we can compare against them.

\begin{figure}[h]
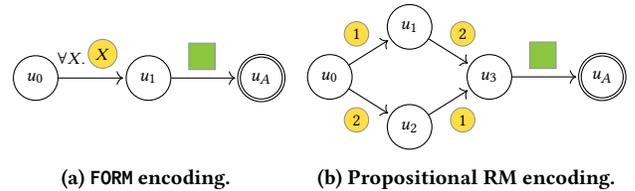

    \centering
    \begin{subfigure}[t]{0.22\textwidth}
        \centering
        \captionsetup{justification=centering}
        \raisebox{18pt}{
            \rmAllYellowFOLHor
        }
        \caption{\FORM{} encoding.}
        \label{fig:all_yellow_form}
        \Description[\FORM{} Encoding of the All Yellow task]{The agent must observe all the yellow checkpoints before going to the goal.}
    \end{subfigure}
    \begin{subfigure}[t]{0.25\textwidth}
        \centering
        \captionsetup{justification=centering}
        \rmAllYellowPropHor
        \caption{Propositional RM encoding.}
        \label{fig:all_yellow_prop}
        \Description[Propositional RM Encoding of the All Yellow task]{The agent must observe all the yellow checkpoints before going to the goal.}
    \end{subfigure}
    \caption{RMs for the \textsc{AllYellow} task.}
    \label{fig:all_yellow_rms}
\end{figure}

\subsubsection*{Objective} We wish to assess the performance impact of learning \FORM{}s as opposed to using a traditional RM learning method \cite{Furelos-Blanco_Law_Jonsson_Broda_Russo_2021}. With fewer states, fewer sub-tasks need to be learnt, which should make the policy learning easier and therefore faster.

\subsubsection*{Results} \figref{fig:all_yellow_rms} presents the \FORM{} and propositional RM encodings of the task and \figref{fig:all_yellow_results} the learning curves of the different approaches. The traditional RL learning algorithm PPO (in \textit{blue}) fails at solving this non-Markovian task since it does not learn history-dependent policies by default. \update{Recurrent PPO (in \textit{purple}) converges rapidly to a high return but fails to find the optimal policy.} On the other hand, the RM-based solutions successfully learn an optimal policy to solve the task. The perfect \FORM{} (\figref{fig:all_yellow_form}) is learnt quickly (policy learning iteration \#3) by our algorithm. Similarly, the perfect propositional RM (\figref{fig:all_yellow_prop}) is learnt at iteration \#4; however, it requires learning more policies since the RM has more states. Our method (in \textit{red}) converges significantly faster than the propositional RM learning method (in \textit{green}), almost on par with the case where the RM is not learnt (in \textit{orange}).

\subsubsection*{\textbf{Task 2 -- \textsc{GreenButOne-NoLava}}}
For this task, the environment is made more complex with the introduction of five randomly placed lava cells $\lavaCell$, ending the episode whenever the agent steps on any of them. The number of green checkpoints $\checkpointEx{green}{}$ is increased from two to three. In this task, the agent must visit \emph{any} green checkpoint $\checkpointEx{green}{}$ different than $\checkpointEx{green}{12}$ (which is also green), then reach the goal $\goalCell$ but without ever stepping on the lava $\lavaCell$.

\begin{figure}[h]
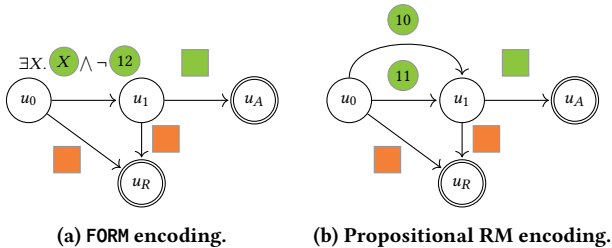

    \begin{subfigure}[t]{0.235\textwidth}
        \centering
        \captionsetup{justification=centering}
        \rmGreenNoLavaFOLHor
        \caption{\FORM{} encoding.}
        \label{fig:green_no_lava_form}
        \Description[\FORM{} Encoding of the Green But One No Lava task]{The agent must visit the green checkpoints 10 or 11 before reaching the goal without stepping on Lava}
    \end{subfigure}
    \begin{subfigure}[t]{0.235\textwidth}
        \centering
        \captionsetup{justification=centering}
        \rmGreenNoLavaPropHor
        \caption{Propositional RM encoding.}
        \label{fig:green_no_lava_prop}
        \Description[RM Encoding of the Green But One No Lava task]{The agent must visit the green checkpoints 10 or 11 before reaching the goal without stepping on Lava}
    \end{subfigure}
    \caption{RMs for the \textsc{GreenButOne-NoLava} task.}
    \label{fig:green_no_lava_rms}
\end{figure}

\subsubsection*{Objective} This experiment intends to show the expressivity of the learnable rules, including complex disjunctions between first-order formulae and ground atoms or propositions, and the ability to learn a \FORM{} with rejecting states.

\subsubsection*{Results} The results are presented in \figref{fig:green_no_lava_results}, where we can see that the Handcrafted \FORM{} baseline (in \emph{orange}) outperforms the other two RM-based approaches converging faster. However, both the Propositional RM-based method (in \emph{green}) and ours (in \emph{red}) achieve a very similar level of performance. In both cases, they manage to correctly learn the RMs illustrated in \figref{fig:green_no_lava_rms}, showing that akin to traditional RM learning methods, our \FORM{}-learning method can induce complex task structures involving both accepting and rejecting states. Both RMs share the same number of states, and thus the policies to train, which explains the similar performance.

\subsubsection*{\textbf{Task 3 -- \textsc{Blue-AllYellow-7}}}
The task consists of visiting \emph{any} blue checkpoint $\checkpointEx{blue}{}$, followed by visiting all yellow checkpoints $\checkpointEx{yellow}{}$, visiting the checkpoint $\checkpointEx{violet}{7}$, and reaching the $\goalCell$ location. The grid in \figref{fig:minigrid_ex} contains two checkpoints for each colour. This task is much more complex than the previous ones since it combines existentially and universally quantified atoms, ground atoms, and propositions over a long sequence of sub-tasks.

\begin{figure}[h]
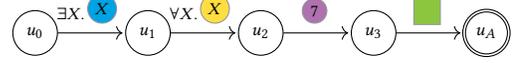

    \begin{subfigure}[t]{0.47\textwidth}
        \centering
        \captionsetup{justification=centering}
        \raisebox{7pt}{
            \rmBlueAllYellowSevenFOLHor
        }
        \Description[\FORM{} encoding the Blue-All Yellow-7 task]{The reward machine encoding the task where the agent has to find a blue checkpoint followed by all the yellow checkpoints followed by the checkpoint with ID 7 to finally reach the goal.}
    \end{subfigure}
    \caption{\FORM{} for the \textsc{Blue-AllYellow-7} task.}
    \label{fig:blue_all_yellow_7_form}
\end{figure}

\subsubsection*{Objective} This experiment aims to demonstrate the scalability of our approach compared to existing RM learning techniques. The expressivity of the first-order language supported by \FORM{s} enables a more compact encoding, addressing the scalability concerns of learning propositional RMs.

\subsubsection*{Results} The results are shown in \figref{fig:blue_all_yellow_7_results}. Our approach manages to learn the \FORM{} encoding the structure of this task (\figref{fig:blue_all_yellow_7_form}), while the propositional RM approach (in \textit{green}) fails to find the correct RM (see \figref{fig:blue_all_yellow_7_rm_prop} in \appendixref{apdx:additional_examples}) and times out due to the size of the RM to learn (two more states and twice as many edges as the \FORM{} encoding). Moreover, our approach (in \textit{red}) reaches the same performance level as that with a handcrafted \FORM{} (in \textit{orange}). Our approach therefore scales to more complex problems whereas existing methods cannot.

\subsection{\FORM{} Transfer}

One of the advantages of \FORM{}s lies in their ability to learn RMs that capture the structure of a task without necessarily being bound to the individual objects of the environments but rather to their properties (e.g., their colour in our case). It makes the RM transferable to new tasks, considerably improving learning efficiency. Our multi-agent formulation lets us also transfer the trained policies to the new task. Depending on the `a priori' knowledge available, we may want to restrict the re-training to the ones that need it (see results below) or re-train all of them (see \appendixref{apdx:transfer}). In both cases, the transferred policies are used as a warm start.

\begin{figure}[h]
    \centering
    \captionsetup{justification=centering}
    \begin{subfigure}[t]{0.49\textwidth}
        \centering
        \includegraphics[width=0.75\linewidth]{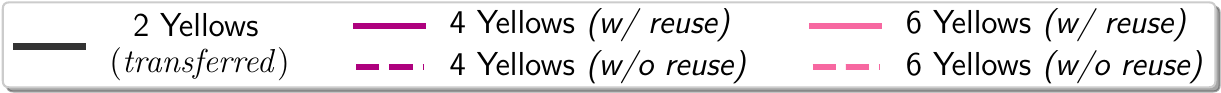}
    \end{subfigure}
    \hfill
    \begin{subfigure}[t]{0.49\textwidth}
        \centering
        \includegraphics[width=0.7\linewidth]{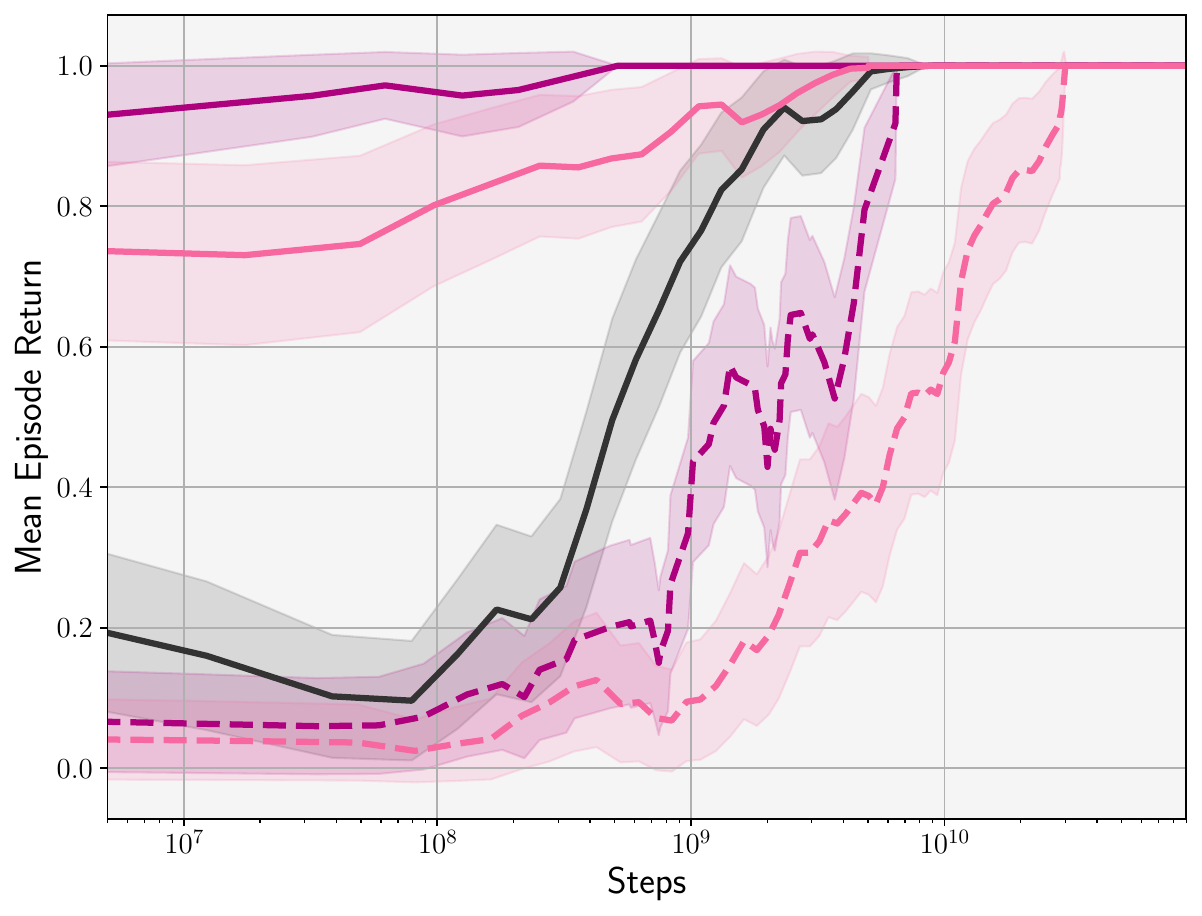}
    \end{subfigure}
    \caption{Transfer of the \FORM{} learnt for the task from \figref{fig:fol_rm_task}, to an environment with more yellow checkpoints \textit{(w/ reuse)} compared against full \FORM{} learning \textit{(w/o reuse)}.}
    \Description{Episode Return Mean transferring the RM and the policies learnt with the task from \figref{fig:fol_rm_task} with 2 yellow checkpoints \textit{(w/ reuse)}, to an environment with 4 and 6 yellow checkpoints. We compare against \FORM{} learning the RM and the policies from scratch \textit{(w/o reuse)}.}
    \label{fig:transfer}
\end{figure}


\subsubsection*{Setup} We reuse the \FORM{} and policies learnt for the \textsc{AllYellow} task (\figref{fig:all_yellow_results}) in two tasks. These tasks increase the number of yellow checkpoints $\checkpointEx{yellow}{}$ to four and six, respectively. We evaluate the change in learning speed between learning from scratch \emph{(w/o reuse)} and reusing the \FORM{} and policies \emph{(w/ reuse)}. 
%
%
The policy associated with the RM state $\uI$ where the agent must ``visit all $\checkpointEx{yellow}{}$'' is retrained, whereas that for going to $\goalCell$ remains unchanged.

\subsubsection*{Objective} This experiment aims to show the transferability of \FORM{}s and the ability to adapt quickly to new tasks. \FORM{}s help abstract the environment settings, while the multi-agent formulation helps the transfer of the policies associated with the sub-tasks.

\subsubsection*{Results} \figref{fig:transfer} shows \textit{(in black)} the performance of the base solution being transferred (two yellow checkpoints) along with the performance of our approach learning the \FORM{} and the policies for four and six yellow checkpoints \textit{(coloured dotted lines)}. We then reuse the transferred \FORM{} and policies to solve those tasks \textit{(coloured solid lines)}. We see a significant speed-up in the learning of the optimal policies in both cases, highlighting the benefits of reusing existing solutions trained on simpler tasks. See \appendixref{apdx:transfer} for more experiments.

\section{Related Work}
\label{sec:rw}

\subsubsection*{\textbf{Exploiting RMs}}
Recent work has also aimed at increasing the expressivity of RMs, e.g.~by introducing counter variables that support richer grammars~\cite{bestercounting}. In parallel, other languages such as \emph{linear temporal logic} have also been used to formulate task specifications~\cite{ijcai2019p840, toro_icarte_multi_tasks_ltl,degiacomo2020transducers,Araki2021TheLO}.  These approaches primarily focus on \emph{exploiting} the \emph{given} task structure. Unlike these methods, our approach tackles the challenge of \emph{learning} the task structure itself.

The exploitation of RMs is often performed in the literature through QRM~\cite{icarte2018using} or CRM~\cite{Icarte_Klassen_Valenzano_McIlraith_2022}. Although both learn policies over  the joint state space $\envStatesSet\times\statesSet$ of the MDP and RM, the former associates an action-value function with each RM state, whereas the latter keeps a single action-value function. These algorithms employ a counterfactual mechanism that enables reusing the experiences for an RM state to update the value associated with other RM states; as such, the underlying RL algorithms are off-policy, e.g.~DDQN~\cite{Icarte_Klassen_Valenzano_McIlraith_2022}, DDPG~\cite{Icarte_Klassen_Valenzano_McIlraith_2022,10.1007/978-3-031-43264-4_6}, and SAC~\cite{10.1007/978-3-031-43264-4_6}. 
\update{QRM and our approach both learn a policy for each RM state; however our approach uses an on-policy algorithm (PPO). \citet{liNoisyRM2024} also use PPO with RMs but learn a single policy conditioned on $\envStatesSet\times\statesSet$.}
%
%
Besides, the $Q$-value function in QRM is updated by using the successor state's value function, creating a dependency between the policies of each state. We, on the other hand, take the approach of using the global reward returned by the RM as a signal for each agent, limiting the dependency between the policies, hence increasing transferability.

Our multi-agent exploitation method aligns with the \emph{options} framework~\cite{sutton1999between}. 
In our formulation, each agent operates an option determining the next action to accomplish a sub-task. In our case, an option terminates when the RM state changes (i.e.,~the option for the next RM state initiates). Unlike the options framework, which usually considers rewards during an option's activation, our approach distributes rewards to all agents that contributed to reaching the current RM state.

\subsubsection*{\textbf{Learning RMs}}

Most RM research has focused on relaxing the assumption that the RM is known `a priori', proposing various methods to learn it dynamically. These methods include discrete optimization~\cite{toro2019learning}, SAT solving~\cite{xu2020joint}, state-merging~\cite{gaon2020reinforcement}, inferring LTL formula from expert data~\cite{baert2024learningLTL}, and inductive logic programming~\cite{Furelos-Blanco_Law_Jonsson_Broda_Russo_2021,ParacNAFCR24}, the latter of which forms the base of our approach. These techniques leverage traces collected through the agent's exploration or provided by an expert to \emph{learn} the RM. All these methods have focused on learning propositional RMs, which are tailored to specific scenarios and thus lack reusability. Additionally, as the task complexity increases, the size of the RMs typically grows exponentially, making them impossible to learn. \update{Finally, more loosely related to our work, other methods learn ASP rules as a proxy for the policy \cite{meli2024learning,baugh2025neuraldnfmt}. However, unlike our work, the ASP encode the decision rule of the policy rather than the structure of the task.}

To address the scalability issues inherent to propositional RMs, recent work has proposed to compose them hierarchically~\cite{furelos2023hierarchies}, learning small RMs that call each other. \update{This approach requires knowing the level of each RM in the hierarchy and learns them bottom-up. Besides, existentially and universally quantified atoms cannot be hierarchically decomposed, making HRM suffer the same problem as standard RM learning approaches in these cases.}
Similarly, work on multi-agent RL decomposes a task across multiple agents, each learning their own RM~\cite{ardon2023cooperative}. While these approaches still focus on propositional RMs, they could be extended to incorporate our contributions.

\section{Conclusion}
\label{sec:concl}

While the literature has actively contributed to improving the exploitation and learning of RMs, existing methods primarily focus on propositional RMs, which are limited in their generalizability and scalability. 
In this paper, we introduced First-Order Reward Machines (\FORM{}s), a novel extension of reward machines (RMs) that leverages first-order logic to enhance expressivity and transferability. 
We propose a learning algorithm that effectively \emph{learns} \FORM{}s from traces, and a multi-agent framework that facilitates efficient policy learning and transfer. 
Experimental results demonstrate that, unlike traditional RM methods, \FORM{}s enhance scalability, enable faster policy and RM learning, and ease task transferability. 
%
Although our approach provides greater expressivity than propositional logic, extending our methodology to include languages like LTL could be an interesting avenue to be able to also learn the structure of temporally extended tasks.

\newpage



\bibliographystyle{ACM-Reference-Format}
\balance
\bibliography{references}


\begin{thebibliography}{42}


\ifx \showCODEN    \undefined \def \showCODEN     #1{\unskip}     \fi
\ifx \showDOI      \undefined \def \showDOI       #1{#1}\fi
\ifx \showISBNx    \undefined \def \showISBNx     #1{\unskip}     \fi
\ifx \showISBNxiii \undefined \def \showISBNxiii  #1{\unskip}     \fi
\ifx \showISSN     \undefined \def \showISSN      #1{\unskip}     \fi
\ifx \showLCCN     \undefined \def \showLCCN      #1{\unskip}     \fi
\ifx \shownote     \undefined \def \shownote      #1{#1}          \fi
\ifx \showarticletitle \undefined \def \showarticletitle #1{#1}   \fi
\ifx \showURL      \undefined \def \showURL       {\relax}        \fi
\providecommand\bibfield[2]{#2}
\providecommand\bibinfo[2]{#2}
\providecommand\natexlab[1]{#1}
\providecommand\showeprint[2][]{arXiv:#2}

\bibitem[\protect\citeauthoryear{Araki, Li, Vodrahalli, DeCastro, Fry, and Rus}{Araki et~al\mbox{.}}{2021}]%
        {Araki2021TheLO}
\bibfield{author}{\bibinfo{person}{Brandon Araki}, \bibinfo{person}{Xiao Li}, \bibinfo{person}{Kiran Vodrahalli}, \bibinfo{person}{Jonathan~A. DeCastro}, \bibinfo{person}{Micah~J. Fry}, {and} \bibinfo{person}{Daniela Rus}.} \bibinfo{year}{2021}\natexlab{}.
\newblock \showarticletitle{{The Logical Options Framework}}. In \bibinfo{booktitle}{\emph{Proceedings of the International Conference on Machine Learning (ICML)}}. \bibinfo{pages}{307--317}.
\newblock


\bibitem[\protect\citeauthoryear{Ardon, Furelos{-}Blanco, and Russo}{Ardon et~al\mbox{.}}{2023}]%
        {ardon2023cooperative}
\bibfield{author}{\bibinfo{person}{Leo Ardon}, \bibinfo{person}{Daniel Furelos{-}Blanco}, {and} \bibinfo{person}{Alessandra Russo}.} \bibinfo{year}{2023}\natexlab{}.
\newblock \showarticletitle{{Learning Reward Machines in Cooperative Multi-Agent Tasks}}. In \bibinfo{booktitle}{\emph{Autonomous Agents and Multiagent Systems. Best and Visionary Papers}}. \bibinfo{pages}{43--59}.
\newblock


\bibitem[\protect\citeauthoryear{Ardon, Vadori, Spooner, Xu, Vann, and Ganesh}{Ardon et~al\mbox{.}}{2021}]%
        {ardon2021towards}
\bibfield{author}{\bibinfo{person}{Leo Ardon}, \bibinfo{person}{Nelson Vadori}, \bibinfo{person}{Thomas Spooner}, \bibinfo{person}{Mengda Xu}, \bibinfo{person}{Jared Vann}, {and} \bibinfo{person}{Sumitra Ganesh}.} \bibinfo{year}{2021}\natexlab{}.
\newblock \showarticletitle{Towards a fully RL-based Market Simulator}. In \bibinfo{booktitle}{\emph{Proceedings of the ACM International Conference on AI in Finance}}. \bibinfo{pages}{1--9}.
\newblock


\bibitem[\protect\citeauthoryear{Baert, Leroux, and Simoens}{Baert et~al\mbox{.}}{2024}]%
        {baert2024learningLTL}
\bibfield{author}{\bibinfo{person}{Mattijs Baert}, \bibinfo{person}{Sam Leroux}, {and} \bibinfo{person}{Pieter Simoens}.} \bibinfo{year}{2024}\natexlab{}.
\newblock \showarticletitle{{Learning Temporal Task Specifications From Demonstrations}}. In \bibinfo{booktitle}{\emph{Proceedings of the International Workshop on Explainable and Transparent {AI} and Multi-Agent Systems (EXTRAAMAS) at the International Conference on Autonomous Agents and Multiagent Systems (AAMAS)}}. \bibinfo{pages}{81--98}.
\newblock


\bibitem[\protect\citeauthoryear{Baugh, Dickens, and Russo}{Baugh et~al\mbox{.}}{2025}]%
        {baugh2025neuraldnfmt}
\bibfield{author}{\bibinfo{person}{Kexin~Gu Baugh}, \bibinfo{person}{Luke Dickens}, {and} \bibinfo{person}{Alessandra Russo}.} \bibinfo{year}{2025}\natexlab{}.
\newblock \showarticletitle{Neural DNF-MT: A Neuro-symbolic Approach for Learning Interpretable and Editable Policies}. In \bibinfo{booktitle}{\emph{Proceedings of the International Conference on Autonomous Agents and Multiagent Systems (AAMAS)}}.
\newblock


\bibitem[\protect\citeauthoryear{Bester, Rosman, James, and Nangue~Tasse}{Bester et~al\mbox{.}}{2024}]%
        {bestercounting}
\bibfield{author}{\bibinfo{person}{Tristan Bester}, \bibinfo{person}{Benjamin Rosman}, \bibinfo{person}{Steven James}, {and} \bibinfo{person}{Geraud Nangue~Tasse}.} \bibinfo{year}{2024}\natexlab{}.
\newblock \showarticletitle{{Counting Reward Automata: Sample Efficient Reinforcement Learning Through The Exploitation of Reward Function Structure}}. In \bibinfo{booktitle}{\emph{Proceedings of the Neuro-Symbolic Learning and Reasoning in the Era of Large Language Models Workshop (NucLeaR) at the AAAI Conference on Artificial Intelligence (AAAI)}}.
\newblock


\bibitem[\protect\citeauthoryear{Camacho, Chen, Sanner, and McIlraith}{Camacho et~al\mbox{.}}{2017}]%
        {camacho2018non}
\bibfield{author}{\bibinfo{person}{Alberto Camacho}, \bibinfo{person}{Oscar Chen}, \bibinfo{person}{Scott Sanner}, {and} \bibinfo{person}{Sheila~A. McIlraith}.} \bibinfo{year}{2017}\natexlab{}.
\newblock \showarticletitle{{Non-Markovian Rewards Expressed in {LTL:} Guiding Search Via Reward Shaping}}. In \bibinfo{booktitle}{\emph{Proceedings of the International Symposium on Combinatorial Search ({SOCS})}}. \bibinfo{pages}{159--160}.
\newblock


\bibitem[\protect\citeauthoryear{Camacho, Toro~Icarte, Klassen, Valenzano, and McIlraith}{Camacho et~al\mbox{.}}{2019}]%
        {ijcai2019p840}
\bibfield{author}{\bibinfo{person}{Alberto Camacho}, \bibinfo{person}{Rodrigo Toro~Icarte}, \bibinfo{person}{Toryn~Q. Klassen}, \bibinfo{person}{Richard Valenzano}, {and} \bibinfo{person}{Sheila~A. McIlraith}.} \bibinfo{year}{2019}\natexlab{}.
\newblock \showarticletitle{{LTL and Beyond: Formal Languages for Reward Function Specification in Reinforcement Learning}}. In \bibinfo{booktitle}{\emph{Proceedings of the International Joint Conference on Artificial Intelligence (IJCAI)}}. \bibinfo{pages}{6065--6073}.
\newblock


\bibitem[\protect\citeauthoryear{De~Giacomo, Favorito, Iocchi, Patrizi, and Ronca}{De~Giacomo et~al\mbox{.}}{2020}]%
        {degiacomo2020transducers}
\bibfield{author}{\bibinfo{person}{Giuseppe De~Giacomo}, \bibinfo{person}{Marco Favorito}, \bibinfo{person}{Luca Iocchi}, \bibinfo{person}{Fabio Patrizi}, {and} \bibinfo{person}{Alessandro Ronca}.} \bibinfo{year}{2020}\natexlab{}.
\newblock \showarticletitle{{Temporal Logic Monitoring Rewards via Transducers}}. In \bibinfo{booktitle}{\emph{{Proceedings of the International Conference on Principles of Knowledge Representation and Reasoning (KR)}}}. \bibinfo{pages}{860--870}.
\newblock


\bibitem[\protect\citeauthoryear{Furelos-Blanco, Law, Jonsson, Broda, and Russo}{Furelos-Blanco et~al\mbox{.}}{2021}]%
        {Furelos-Blanco_Law_Jonsson_Broda_Russo_2021}
\bibfield{author}{\bibinfo{person}{Daniel Furelos-Blanco}, \bibinfo{person}{Mark Law}, \bibinfo{person}{Anders Jonsson}, \bibinfo{person}{Krysia Broda}, {and} \bibinfo{person}{Alessandra Russo}.} \bibinfo{year}{2021}\natexlab{}.
\newblock \showarticletitle{{Induction and Exploitation of Subgoal Automata for Reinforcement Learning}}.
\newblock \bibinfo{journal}{\emph{Journal of Artificial Intelligence Research}}  \bibinfo{volume}{70} (\bibinfo{year}{2021}), \bibinfo{pages}{1031–1116}.
\newblock


\bibitem[\protect\citeauthoryear{Furelos-Blanco, Law, Jonsson, Broda, and Russo}{Furelos-Blanco et~al\mbox{.}}{2023}]%
        {furelos2023hierarchies}
\bibfield{author}{\bibinfo{person}{Daniel Furelos-Blanco}, \bibinfo{person}{Mark Law}, \bibinfo{person}{Anders Jonsson}, \bibinfo{person}{Krysia Broda}, {and} \bibinfo{person}{Alessandra Russo}.} \bibinfo{year}{2023}\natexlab{}.
\newblock \showarticletitle{{Hierarchies of Reward Machines}}. In \bibinfo{booktitle}{\emph{Proceedings of the International Conference on Machine Learning (ICML)}}. \bibinfo{pages}{10494--10541}.
\newblock


\bibitem[\protect\citeauthoryear{Gaon and Brafman}{Gaon and Brafman}{2020}]%
        {gaon2020reinforcement}
\bibfield{author}{\bibinfo{person}{Maor Gaon} {and} \bibinfo{person}{Ronen Brafman}.} \bibinfo{year}{2020}\natexlab{}.
\newblock \showarticletitle{{Reinforcement Learning with Non-Markovian Rewards}}. In \bibinfo{booktitle}{\emph{Proceedings of the AAAI Conference on Artificial Intelligence (AAAI)}}. \bibinfo{pages}{3980--3987}.
\newblock


\bibitem[\protect\citeauthoryear{Gelfond and Kahl}{Gelfond and Kahl}{2014}]%
        {gelfond2014knowledge}
\bibfield{author}{\bibinfo{person}{Michael Gelfond} {and} \bibinfo{person}{Yulia Kahl}.} \bibinfo{year}{2014}\natexlab{}.
\newblock \bibinfo{booktitle}{\emph{Knowledge representation, reasoning, and the design of intelligent agents: The answer-set programming approach}}.
\newblock \bibinfo{publisher}{Cambridge University Press}.
\newblock


\bibitem[\protect\citeauthoryear{Gelfond and Lifschitz}{Gelfond and Lifschitz}{1988}]%
        {GelfondL88}
\bibfield{author}{\bibinfo{person}{Michael Gelfond} {and} \bibinfo{person}{Vladimir Lifschitz}.} \bibinfo{year}{1988}\natexlab{}.
\newblock \showarticletitle{{The Stable Model Semantics for Logic Programming}}. In \bibinfo{booktitle}{\emph{Proceedings of the International Conference and Symposium on Logic Programming (ICLP/SLP)}}. \bibinfo{pages}{1070--1080}.
\newblock


\bibitem[\protect\citeauthoryear{Kendall, Hawke, Janz, Mazur, Reda, Allen, Lam, Bewley, and Shah}{Kendall et~al\mbox{.}}{2019}]%
        {kendall2019learning}
\bibfield{author}{\bibinfo{person}{Alex Kendall}, \bibinfo{person}{Jeffrey Hawke}, \bibinfo{person}{David Janz}, \bibinfo{person}{Przemyslaw Mazur}, \bibinfo{person}{Daniele Reda}, \bibinfo{person}{John{-}Mark Allen}, \bibinfo{person}{Vinh{-}Dieu Lam}, \bibinfo{person}{Alex Bewley}, {and} \bibinfo{person}{Amar Shah}.} \bibinfo{year}{2019}\natexlab{}.
\newblock \showarticletitle{Learning to Drive in a Day}. In \bibinfo{booktitle}{\emph{{Proceedings of the International Conference on Robotics and Automation (ICRA)}}}. \bibinfo{pages}{8248--8254}.
\newblock


\bibitem[\protect\citeauthoryear{Law, Russo, and Broda}{Law et~al\mbox{.}}{2015}]%
        {ILASP_system}
\bibfield{author}{\bibinfo{person}{Mark Law}, \bibinfo{person}{Alessandra Russo}, {and} \bibinfo{person}{Krysia Broda}.} \bibinfo{year}{2015}\natexlab{}.
\newblock \bibinfo{title}{{T}he {ILASP} {S}ystem for {L}earning {A}nswer {S}et {P}rograms}.
\newblock
\newblock
\urldef\tempurl%
\url{https://www.ilasp.com}
\showURL{%
\tempurl}


\bibitem[\protect\citeauthoryear{Law, Russo, and Broda}{Law et~al\mbox{.}}{2016}]%
        {LawRB16}
\bibfield{author}{\bibinfo{person}{Mark Law}, \bibinfo{person}{Alessandra Russo}, {and} \bibinfo{person}{Krysia Broda}.} \bibinfo{year}{2016}\natexlab{}.
\newblock \showarticletitle{{Iterative Learning of Answer Set Programs from Context Dependent Examples}}.
\newblock \bibinfo{journal}{\emph{Theory and Practice of Logic Programming}} \bibinfo{volume}{16}, \bibinfo{number}{5-6} (\bibinfo{year}{2016}), \bibinfo{pages}{834--848}.
\newblock


\bibitem[\protect\citeauthoryear{Law, Russo, and Broda}{Law et~al\mbox{.}}{2018}]%
        {LawRB18}
\bibfield{author}{\bibinfo{person}{Mark Law}, \bibinfo{person}{Alessandra Russo}, {and} \bibinfo{person}{Krysia Broda}.} \bibinfo{year}{2018}\natexlab{}.
\newblock \bibinfo{booktitle}{\emph{{The Meta-program Injection Feature in ILASP}}}.
\newblock \bibinfo{type}{{T}echnical {R}eport}.
\newblock
\urldef\tempurl%
\url{https://www.doc.ic.ac.uk/~ml1909/ILASP/inject.pdf}
\showURL{%
\tempurl}


\bibitem[\protect\citeauthoryear{Li, Chen, Klassen, Vaezipoor, Toro~Icarte, and McIlraith}{Li et~al\mbox{.}}{2024}]%
        {liNoisyRM2024}
\bibfield{author}{\bibinfo{person}{Andrew Li}, \bibinfo{person}{Zizhao Chen}, \bibinfo{person}{Toryn Klassen}, \bibinfo{person}{Pashootan Vaezipoor}, \bibinfo{person}{Rodrigo Toro~Icarte}, {and} \bibinfo{person}{Sheila McIlraith}.} \bibinfo{year}{2024}\natexlab{}.
\newblock \showarticletitle{Reward Machines for Deep RL in Noisy and Uncertain Environments}. In \bibinfo{booktitle}{\emph{Proceedings of the Advances in Neural Information Processing Systems Conference (NeurIPS)}}. \bibinfo{pages}{110341--110368}.
\newblock


\bibitem[\protect\citeauthoryear{Liang, Liaw, Nishihara, Moritz, Fox, Goldberg, Gonzalez, Jordan, and Stoica}{Liang et~al\mbox{.}}{2018}]%
        {pmlr-v80-liang18b}
\bibfield{author}{\bibinfo{person}{Eric Liang}, \bibinfo{person}{Richard Liaw}, \bibinfo{person}{Robert Nishihara}, \bibinfo{person}{Philipp Moritz}, \bibinfo{person}{Roy Fox}, \bibinfo{person}{Ken Goldberg}, \bibinfo{person}{Joseph Gonzalez}, \bibinfo{person}{Michael Jordan}, {and} \bibinfo{person}{Ion Stoica}.} \bibinfo{year}{2018}\natexlab{}.
\newblock \showarticletitle{{{RL}lib: Abstractions for Distributed Reinforcement Learning}}. In \bibinfo{booktitle}{\emph{Proceedings of the International Conference on Machine Learning (ICML)}}. \bibinfo{pages}{3053--3062}.
\newblock


\bibitem[\protect\citeauthoryear{Meli, Castellini, and Farinelli}{Meli et~al\mbox{.}}{2024}]%
        {meli2024learning}
\bibfield{author}{\bibinfo{person}{Daniele Meli}, \bibinfo{person}{Alberto Castellini}, {and} \bibinfo{person}{Alessandro Farinelli}.} \bibinfo{year}{2024}\natexlab{}.
\newblock \showarticletitle{Learning Logic Specifications for Policy Guidance in POMDPs: an Inductive Logic Programming Approach}.
\newblock \bibinfo{journal}{\emph{Journal of Artificial Intelligence Research}}  \bibinfo{volume}{79} (\bibinfo{year}{2024}), \bibinfo{pages}{725--776}.
\newblock


\bibitem[\protect\citeauthoryear{Mnih, Kavukcuoglu, Silver, Rusu, Veness, Bellemare, Graves, Riedmiller, Fidjeland, Ostrovski, Petersen, Beattie, Sadik, Antonoglou, King, Kumaran, Wierstra, Legg, and Hassabis}{Mnih et~al\mbox{.}}{2015}]%
        {mnih2015human}
\bibfield{author}{\bibinfo{person}{Volodymyr Mnih}, \bibinfo{person}{Koray Kavukcuoglu}, \bibinfo{person}{David Silver}, \bibinfo{person}{Andrei~A. Rusu}, \bibinfo{person}{Joel Veness}, \bibinfo{person}{Marc~G. Bellemare}, \bibinfo{person}{Alex Graves}, \bibinfo{person}{Martin~A. Riedmiller}, \bibinfo{person}{Andreas Fidjeland}, \bibinfo{person}{Georg Ostrovski}, \bibinfo{person}{Stig Petersen}, \bibinfo{person}{Charles Beattie}, \bibinfo{person}{Amir Sadik}, \bibinfo{person}{Ioannis Antonoglou}, \bibinfo{person}{Helen King}, \bibinfo{person}{Dharshan Kumaran}, \bibinfo{person}{Daan Wierstra}, \bibinfo{person}{Shane Legg}, {and} \bibinfo{person}{Demis Hassabis}.} \bibinfo{year}{2015}\natexlab{}.
\newblock \showarticletitle{Human-level control through deep reinforcement learning}.
\newblock \bibinfo{journal}{\emph{Nature}} \bibinfo{volume}{518}, \bibinfo{number}{7540} (\bibinfo{year}{2015}), \bibinfo{pages}{529--533}.
\newblock


\bibitem[\protect\citeauthoryear{Para\'{c}, Nodari, Ardon, Furelos-Blanco, Cerutti, and Russo}{Para\'{c} et~al\mbox{.}}{2024}]%
        {ParacNAFCR24}
\bibfield{author}{\bibinfo{person}{Roko Para\'{c}}, \bibinfo{person}{Lorenzo Nodari}, \bibinfo{person}{Leo Ardon}, \bibinfo{person}{Daniel Furelos-Blanco}, \bibinfo{person}{Federico Cerutti}, {and} \bibinfo{person}{Alessandra Russo}.} \bibinfo{year}{2024}\natexlab{}.
\newblock \showarticletitle{{Learning Robust Reward Machines from Noisy Labels}}. In \bibinfo{booktitle}{\emph{Proceedings of the International Conference on Principles of Knowledge Representation and Reasoning (KR)}}. \bibinfo{pages}{909--919}.
\newblock


\bibitem[\protect\citeauthoryear{Paszke, Gross, Massa, Lerer, Bradbury, Chanan, Killeen, Lin, Gimelshein, Antiga, Desmaison, K{\"{o}}pf, Yang, DeVito, Raison, Tejani, Chilamkurthy, Steiner, Fang, Bai, and Chintala}{Paszke et~al\mbox{.}}{2019}]%
        {PaszkeGMLBCKLGA19}
\bibfield{author}{\bibinfo{person}{Adam Paszke}, \bibinfo{person}{Sam Gross}, \bibinfo{person}{Francisco Massa}, \bibinfo{person}{Adam Lerer}, \bibinfo{person}{James Bradbury}, \bibinfo{person}{Gregory Chanan}, \bibinfo{person}{Trevor Killeen}, \bibinfo{person}{Zeming Lin}, \bibinfo{person}{Natalia Gimelshein}, \bibinfo{person}{Luca Antiga}, \bibinfo{person}{Alban Desmaison}, \bibinfo{person}{Andreas K{\"{o}}pf}, \bibinfo{person}{Edward~Z. Yang}, \bibinfo{person}{Zachary DeVito}, \bibinfo{person}{Martin Raison}, \bibinfo{person}{Alykhan Tejani}, \bibinfo{person}{Sasank Chilamkurthy}, \bibinfo{person}{Benoit Steiner}, \bibinfo{person}{Lu Fang}, \bibinfo{person}{Junjie Bai}, {and} \bibinfo{person}{Soumith Chintala}.} \bibinfo{year}{2019}\natexlab{}.
\newblock \showarticletitle{{PyTorch: An Imperative Style, High-Performance Deep Learning Library}}. In \bibinfo{booktitle}{\emph{Proceedings of the Advances in Neural Information Processing Systems Conference (NeurIPS)}}. \bibinfo{pages}{8024--8035}.
\newblock


\bibitem[\protect\citeauthoryear{Sallab, Abdou, Perot, and Yogamani}{Sallab et~al\mbox{.}}{2017}]%
        {sallab2017deep}
\bibfield{author}{\bibinfo{person}{Ahmad~El Sallab}, \bibinfo{person}{Mohammed Abdou}, \bibinfo{person}{Etienne Perot}, {and} \bibinfo{person}{Senthil~Kumar Yogamani}.} \bibinfo{year}{2017}\natexlab{}.
\newblock \showarticletitle{Deep Reinforcement Learning framework for Autonomous Driving}. In \bibinfo{booktitle}{\emph{Autonomous Vehicles and Machines}}. \bibinfo{publisher}{Society for Imaging Science and Technology}, \bibinfo{pages}{70--76}.
\newblock


\bibitem[\protect\citeauthoryear{Schulman, Wolski, Dhariwal, Radford, and Klimov}{Schulman et~al\mbox{.}}{2017}]%
        {schulman2017ppo}
\bibfield{author}{\bibinfo{person}{John Schulman}, \bibinfo{person}{Filip Wolski}, \bibinfo{person}{Prafulla Dhariwal}, \bibinfo{person}{Alec Radford}, {and} \bibinfo{person}{Oleg Klimov}.} \bibinfo{year}{2017}\natexlab{}.
\newblock \bibinfo{title}{{Proximal Policy Optimization Algorithms}}.
\newblock
\newblock
\showeprint[arxiv]{1707.06347}


\bibitem[\protect\citeauthoryear{Senior, Evans, Jumper, Kirkpatrick, Sifre, Green, Qin, Z{\'{\i}}dek, Nelson, Bridgland, Penedones, Petersen, Simonyan, Crossan, Kohli, Jones, Silver, Kavukcuoglu, and Hassabis}{Senior et~al\mbox{.}}{2020}]%
        {senior2020improved}
\bibfield{author}{\bibinfo{person}{Andrew~W. Senior}, \bibinfo{person}{Richard Evans}, \bibinfo{person}{John Jumper}, \bibinfo{person}{James Kirkpatrick}, \bibinfo{person}{Laurent Sifre}, \bibinfo{person}{Tim Green}, \bibinfo{person}{Chongli Qin}, \bibinfo{person}{Augustin Z{\'{\i}}dek}, \bibinfo{person}{Alexander W.~R. Nelson}, \bibinfo{person}{Alex Bridgland}, \bibinfo{person}{Hugo Penedones}, \bibinfo{person}{Stig Petersen}, \bibinfo{person}{Karen Simonyan}, \bibinfo{person}{Steve Crossan}, \bibinfo{person}{Pushmeet Kohli}, \bibinfo{person}{David~T. Jones}, \bibinfo{person}{David Silver}, \bibinfo{person}{Koray Kavukcuoglu}, {and} \bibinfo{person}{Demis Hassabis}.} \bibinfo{year}{2020}\natexlab{}.
\newblock \showarticletitle{Improved protein structure prediction using potentials from deep learning}.
\newblock \bibinfo{journal}{\emph{Nature}} \bibinfo{volume}{577}, \bibinfo{number}{7792} (\bibinfo{year}{2020}), \bibinfo{pages}{706--710}.
\newblock


\bibitem[\protect\citeauthoryear{Silver, Huang, Maddison, Guez, Sifre, van~den Driessche, Schrittwieser, Antonoglou, Panneershelvam, Lanctot, Dieleman, Grewe, Nham, Kalchbrenner, Sutskever, Lillicrap, Leach, Kavukcuoglu, Graepel, and Hassabis}{Silver et~al\mbox{.}}{2016}]%
        {silver2016mastering}
\bibfield{author}{\bibinfo{person}{David Silver}, \bibinfo{person}{Aja Huang}, \bibinfo{person}{Chris~J. Maddison}, \bibinfo{person}{Arthur Guez}, \bibinfo{person}{Laurent Sifre}, \bibinfo{person}{George van~den Driessche}, \bibinfo{person}{Julian Schrittwieser}, \bibinfo{person}{Ioannis Antonoglou}, \bibinfo{person}{Vedavyas Panneershelvam}, \bibinfo{person}{Marc Lanctot}, \bibinfo{person}{Sander Dieleman}, \bibinfo{person}{Dominik Grewe}, \bibinfo{person}{John Nham}, \bibinfo{person}{Nal Kalchbrenner}, \bibinfo{person}{Ilya Sutskever}, \bibinfo{person}{Timothy~P. Lillicrap}, \bibinfo{person}{Madeleine Leach}, \bibinfo{person}{Koray Kavukcuoglu}, \bibinfo{person}{Thore Graepel}, {and} \bibinfo{person}{Demis Hassabis}.} \bibinfo{year}{2016}\natexlab{}.
\newblock \showarticletitle{Mastering the game of Go with deep neural networks and tree search}.
\newblock \bibinfo{journal}{\emph{Nature}} \bibinfo{volume}{529}, \bibinfo{number}{7587} (\bibinfo{year}{2016}), \bibinfo{pages}{484--489}.
\newblock


\bibitem[\protect\citeauthoryear{Spooner, Fearnley, Savani, and Koukorinis}{Spooner et~al\mbox{.}}{2018}]%
        {spooner2018marketmaking}
\bibfield{author}{\bibinfo{person}{Thomas Spooner}, \bibinfo{person}{John Fearnley}, \bibinfo{person}{Rahul Savani}, {and} \bibinfo{person}{Andreas Koukorinis}.} \bibinfo{year}{2018}\natexlab{}.
\newblock \showarticletitle{Market Making via Reinforcement Learning}. In \bibinfo{booktitle}{\emph{Proceedings of the International Conference on Autonomous Agents and MultiAgent Systems (AAMAS)}}. \bibinfo{pages}{434–442}.
\newblock


\bibitem[\protect\citeauthoryear{Stokes, Yang, Swanson, Jin, Cubillos-Ruiz, Donghia, MacNair, French, Carfrae, Bloom-Ackermann, et~al\mbox{.}}{Stokes et~al\mbox{.}}{2020}]%
        {stokes2020deep}
\bibfield{author}{\bibinfo{person}{Jonathan~M Stokes}, \bibinfo{person}{Kevin Yang}, \bibinfo{person}{Kyle Swanson}, \bibinfo{person}{Wengong Jin}, \bibinfo{person}{Andres Cubillos-Ruiz}, \bibinfo{person}{Nina~M Donghia}, \bibinfo{person}{Craig~R MacNair}, \bibinfo{person}{Shawn French}, \bibinfo{person}{Lindsey~A Carfrae}, \bibinfo{person}{Zohar Bloom-Ackermann}, {et~al\mbox{.}}} \bibinfo{year}{2020}\natexlab{}.
\newblock \showarticletitle{A deep learning approach to antibiotic discovery}.
\newblock \bibinfo{journal}{\emph{Cell}} \bibinfo{volume}{180}, \bibinfo{number}{4} (\bibinfo{year}{2020}), \bibinfo{pages}{688--702}.
\newblock


\bibitem[\protect\citeauthoryear{Sun and Lesp{\'{e}}rance}{Sun and Lesp{\'{e}}rance}{2023}]%
        {10.1007/978-3-031-43264-4_6}
\bibfield{author}{\bibinfo{person}{Haolin Sun} {and} \bibinfo{person}{Yves Lesp{\'{e}}rance}.} \bibinfo{year}{2023}\natexlab{}.
\newblock \showarticletitle{{Exploiting Reward Machines with Deep Reinforcement Learning in Continuous Action Domains}}. In \bibinfo{booktitle}{\emph{Proceedings of the European Conference on Multi-Agent Systems (EUMAS)}}. \bibinfo{pages}{83--99}.
\newblock


\bibitem[\protect\citeauthoryear{Sutton and Barto}{Sutton and Barto}{2018}]%
        {SuttonB18}
\bibfield{author}{\bibinfo{person}{Richard~S. Sutton} {and} \bibinfo{person}{Andrew~G. Barto}.} \bibinfo{year}{2018}\natexlab{}.
\newblock \bibinfo{booktitle}{\emph{{Reinforcement Learning: An Introduction}}}.
\newblock \bibinfo{publisher}{{MIT} Press}.
\newblock


\bibitem[\protect\citeauthoryear{Sutton, McAllester, Singh, and Mansour}{Sutton et~al\mbox{.}}{1999a}]%
        {SuttonMSM99}
\bibfield{author}{\bibinfo{person}{Richard~S. Sutton}, \bibinfo{person}{David~A. McAllester}, \bibinfo{person}{Satinder Singh}, {and} \bibinfo{person}{Yishay Mansour}.} \bibinfo{year}{1999}\natexlab{a}.
\newblock \showarticletitle{{Policy Gradient Methods for Reinforcement Learning with Function Approximation}}. In \bibinfo{booktitle}{\emph{Proceedings of the Advances in Neural Information Processing Systems Conference (NeurIPS)}}. \bibinfo{pages}{1057--1063}.
\newblock


\bibitem[\protect\citeauthoryear{Sutton, Precup, and Singh}{Sutton et~al\mbox{.}}{1999b}]%
        {sutton1999between}
\bibfield{author}{\bibinfo{person}{Richard~S Sutton}, \bibinfo{person}{Doina Precup}, {and} \bibinfo{person}{Satinder Singh}.} \bibinfo{year}{1999}\natexlab{b}.
\newblock \showarticletitle{Between MDPs and semi-MDPs: A framework for temporal abstraction in reinforcement learning}.
\newblock \bibinfo{journal}{\emph{Artificial Intelligence}}  \bibinfo{volume}{112} (\bibinfo{year}{1999}), \bibinfo{pages}{181--211}.
\newblock


\bibitem[\protect\citeauthoryear{Toro~Icarte, Klassen, Valenzano, and McIlraith}{Toro~Icarte et~al\mbox{.}}{2018a}]%
        {toro_icarte_multi_tasks_ltl}
\bibfield{author}{\bibinfo{person}{Rodrigo Toro~Icarte}, \bibinfo{person}{Toryn~Q. Klassen}, \bibinfo{person}{Richard Valenzano}, {and} \bibinfo{person}{Sheila~A. McIlraith}.} \bibinfo{year}{2018}\natexlab{a}.
\newblock \showarticletitle{{Teaching Multiple Tasks to an RL Agent using LTL}}. In \bibinfo{booktitle}{\emph{Proceedings of the International Conference on Autonomous Agents and Multi-Agent Systems (AAMAS)}}. \bibinfo{pages}{452–461}.
\newblock


\bibitem[\protect\citeauthoryear{Toro~Icarte, Klassen, Valenzano, and McIlraith}{Toro~Icarte et~al\mbox{.}}{2018b}]%
        {icarte2018using}
\bibfield{author}{\bibinfo{person}{Rodrigo Toro~Icarte}, \bibinfo{person}{Toryn~Q. Klassen}, \bibinfo{person}{Richard Valenzano}, {and} \bibinfo{person}{Sheila~A. McIlraith}.} \bibinfo{year}{2018}\natexlab{b}.
\newblock \showarticletitle{{Using Reward Machines for High-Level Task Specification and Decomposition in Reinforcement Learning}}. In \bibinfo{booktitle}{\emph{Proceedings of the International Conference on Machine Learning (ICML)}}. \bibinfo{pages}{2112--2121}.
\newblock


\bibitem[\protect\citeauthoryear{Toro~Icarte, Klassen, Valenzano, and McIlraith}{Toro~Icarte et~al\mbox{.}}{2022}]%
        {Icarte_Klassen_Valenzano_McIlraith_2022}
\bibfield{author}{\bibinfo{person}{Rodrigo Toro~Icarte}, \bibinfo{person}{Toryn~Q. Klassen}, \bibinfo{person}{Richard Valenzano}, {and} \bibinfo{person}{Sheila~A. McIlraith}.} \bibinfo{year}{2022}\natexlab{}.
\newblock \showarticletitle{{Reward Machines: Exploiting Reward Function Structure in Reinforcement Learning}}.
\newblock \bibinfo{journal}{\emph{Journal of Artificial Intelligence Research}}  \bibinfo{volume}{73} (\bibinfo{year}{2022}), \bibinfo{pages}{173–208}.
\newblock


\bibitem[\protect\citeauthoryear{Toro~Icarte, Waldie, Klassen, Valenzano, Castro, and McIlraith}{Toro~Icarte et~al\mbox{.}}{2019}]%
        {toro2019learning}
\bibfield{author}{\bibinfo{person}{Rodrigo Toro~Icarte}, \bibinfo{person}{Ethan Waldie}, \bibinfo{person}{Toryn Klassen}, \bibinfo{person}{Rick Valenzano}, \bibinfo{person}{Margarita Castro}, {and} \bibinfo{person}{Sheila McIlraith}.} \bibinfo{year}{2019}\natexlab{}.
\newblock \showarticletitle{{Learning Reward Machines for Partially Observable Reinforcement Learning}}. In \bibinfo{booktitle}{\emph{Proceedings of the Advances in Neural Information Processing Systems Conference (NeurIPS)}}.
\newblock


\bibitem[\protect\citeauthoryear{Vinyals, Babuschkin, Czarnecki, Mathieu, Dudzik, Chung, Choi, Powell, Ewalds, Georgiev, Oh, Horgan, Kroiss, Danihelka, Huang, Sifre, Cai, Agapiou, Jaderberg, Vezhnevets, Leblond, Pohlen, Dalibard, Budden, Sulsky, Molloy, Paine, G{\"{u}}l{\c{c}}ehre, Wang, Pfaff, Wu, Ring, Yogatama, W{\"{u}}nsch, McKinney, Smith, Schaul, Lillicrap, Kavukcuoglu, Hassabis, Apps, and Silver}{Vinyals et~al\mbox{.}}{2019}]%
        {vinyals2019grandmaster}
\bibfield{author}{\bibinfo{person}{Oriol Vinyals}, \bibinfo{person}{Igor Babuschkin}, \bibinfo{person}{Wojciech~M. Czarnecki}, \bibinfo{person}{Micha{\"{e}}l Mathieu}, \bibinfo{person}{Andrew Dudzik}, \bibinfo{person}{Junyoung Chung}, \bibinfo{person}{David~H. Choi}, \bibinfo{person}{Richard Powell}, \bibinfo{person}{Timo Ewalds}, \bibinfo{person}{Petko Georgiev}, \bibinfo{person}{Junhyuk Oh}, \bibinfo{person}{Dan Horgan}, \bibinfo{person}{Manuel Kroiss}, \bibinfo{person}{Ivo Danihelka}, \bibinfo{person}{Aja Huang}, \bibinfo{person}{Laurent Sifre}, \bibinfo{person}{Trevor Cai}, \bibinfo{person}{John~P. Agapiou}, \bibinfo{person}{Max Jaderberg}, \bibinfo{person}{Alexander~Sasha Vezhnevets}, \bibinfo{person}{R{\'{e}}mi Leblond}, \bibinfo{person}{Tobias Pohlen}, \bibinfo{person}{Valentin Dalibard}, \bibinfo{person}{David Budden}, \bibinfo{person}{Yury Sulsky}, \bibinfo{person}{James Molloy}, \bibinfo{person}{Tom~Le Paine}, \bibinfo{person}{{\c{C}}aglar G{\"{u}}l{\c{c}}ehre}, \bibinfo{person}{Ziyu Wang},
  \bibinfo{person}{Tobias Pfaff}, \bibinfo{person}{Yuhuai Wu}, \bibinfo{person}{Roman Ring}, \bibinfo{person}{Dani Yogatama}, \bibinfo{person}{Dario W{\"{u}}nsch}, \bibinfo{person}{Katrina McKinney}, \bibinfo{person}{Oliver Smith}, \bibinfo{person}{Tom Schaul}, \bibinfo{person}{Timothy~P. Lillicrap}, \bibinfo{person}{Koray Kavukcuoglu}, \bibinfo{person}{Demis Hassabis}, \bibinfo{person}{Chris Apps}, {and} \bibinfo{person}{David Silver}.} \bibinfo{year}{2019}\natexlab{}.
\newblock \showarticletitle{Grandmaster level in StarCraft {II} using multi-agent reinforcement learning}.
\newblock \bibinfo{journal}{\emph{Nature}} \bibinfo{volume}{575}, \bibinfo{number}{7782} (\bibinfo{year}{2019}), \bibinfo{pages}{350--354}.
\newblock


\bibitem[\protect\citeauthoryear{Wurman, Barrett, Kawamoto, MacGlashan, Subramanian, Walsh, Capobianco, Devlic, Eckert, Fuchs, Gilpin, Khandelwal, Kompella, Lin, MacAlpine, Oller, Seno, Sherstan, Thomure, Aghabozorgi, Barrett, Douglas, Whitehead, D{\"{u}}rr, Stone, Spranger, and Kitano}{Wurman et~al\mbox{.}}{2022}]%
        {wurman2022outracing}
\bibfield{author}{\bibinfo{person}{Peter~R. Wurman}, \bibinfo{person}{Samuel Barrett}, \bibinfo{person}{Kenta Kawamoto}, \bibinfo{person}{James MacGlashan}, \bibinfo{person}{Kaushik Subramanian}, \bibinfo{person}{Thomas~J. Walsh}, \bibinfo{person}{Roberto Capobianco}, \bibinfo{person}{Alisa Devlic}, \bibinfo{person}{Franziska Eckert}, \bibinfo{person}{Florian Fuchs}, \bibinfo{person}{Leilani Gilpin}, \bibinfo{person}{Piyush Khandelwal}, \bibinfo{person}{Varun~Raj Kompella}, \bibinfo{person}{HaoChih Lin}, \bibinfo{person}{Patrick MacAlpine}, \bibinfo{person}{Declan Oller}, \bibinfo{person}{Takuma Seno}, \bibinfo{person}{Craig Sherstan}, \bibinfo{person}{Michael~D. Thomure}, \bibinfo{person}{Houmehr Aghabozorgi}, \bibinfo{person}{Leon Barrett}, \bibinfo{person}{Rory Douglas}, \bibinfo{person}{Dion Whitehead}, \bibinfo{person}{Peter D{\"{u}}rr}, \bibinfo{person}{Peter Stone}, \bibinfo{person}{Michael Spranger}, {and} \bibinfo{person}{Hiroaki Kitano}.} \bibinfo{year}{2022}\natexlab{}.
\newblock \showarticletitle{Outracing champion Gran Turismo drivers with deep reinforcement learning}.
\newblock \bibinfo{journal}{\emph{Nature}} \bibinfo{volume}{602}, \bibinfo{number}{7896} (\bibinfo{year}{2022}), \bibinfo{pages}{223--228}.
\newblock


\bibitem[\protect\citeauthoryear{Xu, Gavran, Ahmad, Majumdar, Neider, Topcu, and Wu}{Xu et~al\mbox{.}}{2020}]%
        {xu2020joint}
\bibfield{author}{\bibinfo{person}{Zhe Xu}, \bibinfo{person}{Ivan Gavran}, \bibinfo{person}{Yousef Ahmad}, \bibinfo{person}{Rupak Majumdar}, \bibinfo{person}{Daniel Neider}, \bibinfo{person}{Ufuk Topcu}, {and} \bibinfo{person}{Bo Wu}.} \bibinfo{year}{2020}\natexlab{}.
\newblock \showarticletitle{{Joint Inference of Reward Machines and Policies for Reinforcement Learning}}. In \bibinfo{booktitle}{\emph{Proceedings of the International Conference on Automated Planning and Scheduling (ICAPS)}}. \bibinfo{pages}{590--598}.
\newblock


\bibitem[\protect\citeauthoryear{Yu, Velu, Vinitsky, Gao, Wang, Bayen, and Wu}{Yu et~al\mbox{.}}{2022}]%
        {yu2022surprising}
\bibfield{author}{\bibinfo{person}{Chao Yu}, \bibinfo{person}{Akash Velu}, \bibinfo{person}{Eugene Vinitsky}, \bibinfo{person}{Jiaxuan Gao}, \bibinfo{person}{Yu Wang}, \bibinfo{person}{Alexandre~M. Bayen}, {and} \bibinfo{person}{Yi Wu}.} \bibinfo{year}{2022}\natexlab{}.
\newblock \showarticletitle{{The Surprising Effectiveness of {PPO} in Cooperative Multi-Agent Games}}. In \bibinfo{booktitle}{\emph{Proceedings of the Advances in Neural Information Processing Conference (NeurIPS)}}.
\newblock


\end{thebibliography}


\section*{Supplementary Material}
\appendix

\section{Reward Machines}\label{apdx:additional_examples}
\begin{figure}[h]
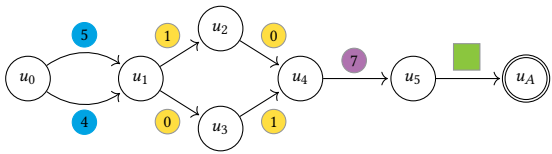

    \centering
    \captionsetup{justification=centering}
    \captionsetup[subfigure]{justification=centering}
    \begin{subfigure}{0.49\textwidth}
        \centering
        \rmBlueAllYellowSevenPropHor
    \end{subfigure}
    \caption{Propositional RM encoding the \textsc{Blue-AllYellow-7} task.}
    \Description{Propositional RM encoding the ``Blue-All Yellow-7'' task.}
    \label{fig:blue_all_yellow_7_rm_prop}
\end{figure}

In this section, we present some reward machines (RMs) omitted in the main paper. \figref{fig:universal_motivation_3} exemplifies the equivalence between traditional and first-order RMs originally shown in \figref{fig:universal_motivation} but extended to a case with three objects. \figref{fig:blue_all_yellow_7_rm_prop} shows the propositional RM for the \textsc{Blue-AllYellow-7} task, whose first-order RM is shown in \figref{fig:blue_all_yellow_7_form}.

\universalMotivationFigThree{h}

\section{\FORM{} Learning Process}
\update{We detail in this section the RM learning algorithm from \cite{Furelos-Blanco_Law_Jonsson_Broda_Russo_2021} used as the base of our approach to learn \FORM{}s. There are two types of traces collected by the agent: (i) \emph{goal} traces, which achieve the task's goal (e.g., visiting all $\checkpointEx{yellow}{}$ followed by visiting $\goalCell$), and (ii) \emph{incomplete} traces (i.e. any trace that is not a goal trace). ILASP must learn a transition function such that goal traces reach the accepting state, and incomplete traces do not reach the accepting state. The trace type (\emph{goal}, \emph{incomplete}) is determined by the environment according to whether the task's goal is reached or not.}

\update{We start from a dummy \FORM{} consisting of an initial state without outgoing edges. The learning of new \FORM{}s is driven by counterexamples, i.e. goal traces that do not reach the accepting state in the current \FORM{}, or incomplete traces that reach the accepting state. The agent cannot initially exploit a valuable \FORM{}, so it acts randomly. The learning of the first \FORM{} is triggered when a goal trace is observed; e.g. for \textsc{AllYellow}, a goal trace always contains a subsequence of yellow checkpoints followed by the goal cell. A goal trace will never contain just a fraction of yellow cells because the environment will not label the trace as a goal trace in such a case. The first learnt \FORM{} will probably not be the target one, e.g. it may just contain an edge labeled with a yellow checkpoint or the goal cell. In the following steps, the agent will collect counterexamples that refute the hypothesis that the current \FORM{} is the target one, and drive the agent towards learning the target one. Theorem 7.1 from \cite{Furelos-Blanco_Law_Jonsson_Broda_Russo_2021} shows that if the target \FORM{} is in the hypothesis space, it will be found with a finite number of counterexamples. We aim to learn a \FORM{} with the fewest number of states and transitions, so the ILASP learning task is such that a solution with a minimal number of rules selected from the hypothesis space will be preferred. Thus, events irrelevant to the target task (e.g., blue circles for \textsc{AllYellow}) will either be omitted from the \FORM{} or will eventually be discarded through counterexamples if they ever appear in a candidate \FORM{}.}

\begin{table}[h]
    \caption{Hyperparameters used  for Handcrafted RM, Propositional RM Learning and \FORM{} Learning.}
    \label{tab:hyperparameters_multi_policy}
    \centering
    \begin{tabular}{@{}ll@{}}
        \toprule
        \textbf{Hyperparameter} & \textbf{Value} \\ \midrule
        Gamma                   & 0.999 \\
        Learning Rate           & 7e-4           \\
        Train Batch Size              & 16384              \\
        SGD Mini Batch Size        & 4096              \\
        Number SGD Iteration & 20 \\
        Value Function Loss Coeff.  & 0.5 \\
        Entropy Coeff. & 0.01 \\
        Clip ratio & 0.15 \\
        Value Function Clip ratio & 0.1 \\
        Gradient Clip & 0.7 \\
        KL Target & 0.15 \\
        Optimizer               & Adam            \\
        GAE & True \\
        Lambda & 0.95 \\
        \bottomrule
        Number Hidden Layers & 2 \\
        Hidden Dimension & 64 \\
        Activation Function & $\tanh$ \\
        CNN Conv Filters (Channel, Kernel, Stride) & [16, [2, 2], 1], \\
                & [32, [2, 2], 1], \\
                & [32, [2, 2], 1] \\
        CNN Max Pooling (Filter, Stride) & [2, 2], \\ 
        & None, \\ 
        & None \\
        \bottomrule
    \end{tabular}
\end{table}

\section{Experimental Details}\label{apdx:experiment_details}

\update{The code is available at \url{https://github.com/leoardon/form}.} The experiments are run using the RLlib framework \cite{pmlr-v80-liang18b}. All experiments were run on a \textsc{c6a.12xlarge} AWS machine, parallelizing the runs for 12 different seeds. An episode is automatically terminated when the agent has accomplished the task 
%
%
or when it has reached the maximum number of time-steps ($3000$), whichever comes first. We set a timeout of $1h$ for \texttt{ILASP} to return a solution. We apply early-stopping of the training whenever the mean return falls by more than $10\%$ of its current value after crossing the limit of $70\%$ of the maximum possible return.

\begin{table}[t]
    \caption{Hyperparameters used for single policy learning.}
    \label{tab:hyperparameters_single_policy}
    \centering
    \begin{tabular}{@{}ll@{}}
        \toprule
        \textbf{Hyperparameter} & \textbf{Value} \\ \midrule
        Gamma                   & 0.999 \\
        Learning Rate           & 7e-4           \\
        Train Batch Size              & 49152              \\
        SGD Mini Batch Size        & 256              \\
        Number SGD Iteration & 20 \\
        Value Function Loss Coeff.  & 0.5 \\
        Entropy Coeff. & 0.01 \\
        Clip ratio & 0.15 \\
        Value Function Clip ratio & 0.1 \\
        Gradient Clip & 0.7 \\
        KL Target & 0.15 \\
        Optimizer               & Adam            \\
        GAE & True \\
        Lambda & 0.95 \\
        \bottomrule
        Number Hidden Layers & 2 \\
        Hidden Dimension & 64 \\
        Activation Function & Tanh \\
        CNN Conv Filters (Channel, Kernel, Stride) & [16, [2, 2], 1], \\
                & [32, [2, 2], 1], \\
                & [32, [2, 2], 1] \\
        CNN Max Pooling (Filter, Stride) & [2, 2], \\ 
        & None, \\ 
        & None \\
        \bottomrule
        Number LSTM Layers & 1 \\
        LSTM Hidden Dimension & 256 \\
        \bottomrule
    \end{tabular}
\end{table}

\subsubsection*{Policy Learning}\label{apdx:model_details}
We use RLlib's PyTorch~\cite{PaszkeGMLBCKLGA19} implementation of PPO. Default weight initializations are employed. A convolutional neural network (CNN) processes the image of the environment, and multi-layer perceptrons process the buffer and the RM state. The outputs of the CNN and the MLPs are concatenated and fed to the policy network. \update{We use the reward shaping technique from \cite{camacho2018non} using the distance to the closest accepting state of the RM as potential function, for all the experiments with an RM (Propositional or First-Order).} In the single policy experiments (see Appendix~\ref{apdx:transfer}), there is only the CNN and one MLP (i.e., that for the RM state is processed). The deep layers of the value function network are shared between the actor and the critic. The hyperparameters for the multi-agent and single policy settings are presented in Tables~\ref{tab:hyperparameters_multi_policy} and \ref{tab:hyperparameters_single_policy}, respectively.

\section{ILASP Learning Timings}

\update{\tblref{tab:ilasp_learning_timings} shows the timings associated with the last ILASP task (i.e. the one that learnt the correct RM) for the baseline method to learn propositional RM and for our \FORM{} learning approach. In general, since \FORM{}s are more compact (i.e. they have fewer states and edges), they can be learnt faster. The performance is comparable when the number of states is the same, such as in the \textsc{GreenButOne-NoLava} task.}

\begin{table}[b]
    \caption{ILASP Learning Timings between Propositional RM learning and \FORM{} learning.}
    \label{tab:ilasp_learning_timings}
    \centering
    \begin{tabular}{c|cc}
        \toprule
        & \textbf{Propositional RM} & \textbf{\FORM{}} \textit{(ours)} \\ 
        \midrule
        \textbf{\textsc{AllYellow}} & 16.3s & 1.0s \\ 
        \textbf{\textsc{GreenButOne-NoLava}} & 0.6s & 0.5s \\
        \textbf{\textsc{Blue-AllYellow-7}} & Timed Out & 12.8s \\
        \bottomrule
    \end{tabular}
\end{table}

\section{Determinism ASP Encoding}\label{apdx:determinism_encoding}
In this section, we describe the ASP encoding of the determinism-related constraints, whose intuition is outlined in \secref{sec:form_learning}. These rules enforce mutual exclusivity between the formulae labelling transitions from a given state to two different states.

The ASP representation of the formulae labelling the edges is done using rules; however, defining constraints over rules is not straightforward. To address this problem, we associate rules in the hypothesis spaces with \emph{facts}, over which defining the mutual exclusivity constraints is easy. The mapping from rules to facts is:
\begin{align*}
\footnotesize
\left\lbrace\begin{array}{@{}l|l@{}}
\texttt{pos(X,Y,E,obs(o)).} & \bar{\transitionFormulaFunction}\texttt{(X,Y,E,T)} \allowbreak \codeif \texttt{not~obs(o,T),step(T).}\\
\texttt{neg(X,Y,E,obs(o)).} & \bar{\transitionFormulaFunction}\texttt{(X,Y,E,T)} \codeif \texttt{obs(o,T),step(T).}\\
\texttt{pos(X,Y,E,e\_pred(p)).} & \bar{\transitionFormulaFunction}\texttt{(X,Y,E,T)} \codeif \texttt{not~e\_pred(p,T),step(T).}\\
\texttt{neg(X,Y,E,e\_pred(p)).}& \bar{\transitionFormulaFunction}\texttt{(X,Y,E,T)} \codeif \texttt{e\_pred(p,T),step(T).}\\
\texttt{pos(X,Y,E,a\_pred(p)).} & \bar{\transitionFormulaFunction}\texttt{(X,Y,E,T)} \codeif \texttt{not~a\_pred(p,T),step(T).}\\
\texttt{neg(X,Y,E,a\_pred(p)).} & \bar{\transitionFormulaFunction}\texttt{(X,Y,E,T)} \codeif \texttt{a\_pred(p,T),step(T).}
\end{array}\right\rbrace,
\end{align*}
where $\texttt{pos}$ (resp.~\texttt{neg}) facts denote that a proposition or quantified atom appears positively (resp.~\texttt{negatively}) in the transition. Remember that the actual transition is the negation of $\bar{\transitionFormulaFunction}$, hence why the \texttt{not} are mapped into \texttt{pos}. The \texttt{ILASP} system enables performing the mapping at learning time through meta-program injection~\cite{LawRB18}; therefore, we can express the mutual exclusivity constraints in terms of \texttt{pos} and \texttt{neg} facts.

Before defining the constraint, we need to map quantified atoms into their respective sets of ground atoms, as described in \secref{sec:form_learning}.  \lstref{lst:equivalence} contains the ASP program that performs the mapping.

\begin{example}
Given the grid from \figref{fig:minigrid_ex}, the mapping for the formulae $\neg \exists X. \checkpointEx{blue}{X}$ labelling a transition from state \texttt{X} to state \texttt{Y} using the edge with id \texttt{E} is:
\begin{gather*}
\small
    \left\lbrace\begin{array}{@{}l|l@{}}
        \begin{array}{l}
             \texttt{neg(X,Y,E,obs(}\checkpointEx{blue}{4}))~\land  \\
             \texttt{neg(X,Y,E,obs(}\checkpointEx{blue}{5})) 
        \end{array} &
    \texttt{neg(X,Y,E,e\_pred(}\checkpointEx{blue}{}))
    \end{array}
    \right\rbrace.
\end{gather*}
The mapping for the formula $\exists X. \checkpointEx{blue}{X}$ is analogous:
\begin{gather*}
\small
    \left\lbrace\begin{array}{@{}l|l@{}}
        \begin{array}{l}
             \texttt{pos(X,Y,E,obs(}\checkpointEx{blue}{4}))~\lor \\
             \texttt{pos(X,Y,E,obs(}\checkpointEx{blue}{5}))
        \end{array} & 
        \texttt{pos(X,Y,E,e\_pred(}\checkpointEx{blue}{}))
    \end{array}
    \right\rbrace.
\end{gather*}
\end{example}

\begin{listing}[h]%
\caption{Mapping from Quantified Atoms to Ground Atoms}%
\lstset{numberstyle=\tiny, escapeinside={(*@}{@*)}}
\label{lst:equivalence}%
\begin{lstlisting}
%%%% Assuming the facts:
%%%% "(*@\textcolor{blue}{P}@*)((*@\textcolor{red}{o1}@*)). (*@\textcolor{blue}{P}@*)((*@\textcolor{red}{o2}@*))."

pos(X, Y, E, obs((*@\textcolor{blue}{P}@*)((*@\textcolor{red}{o1}@*)))) :- pos(X, Y, E, a_pred((*@\textcolor{blue}{P}@*))).
pos(X, Y, E, obs((*@\textcolor{blue}{P}@*)((*@\textcolor{red}{o2}@*)))) :- pos(X, Y, E, a_pred((*@\textcolor{blue}{P}@*))).

neg(X, Y, E, obs((*@\textcolor{blue}{P}@*)((*@\textcolor{red}{o1}@*)))) :- neg(X, Y, E, e_pred((*@\textcolor{blue}{P}@*))).
neg(X, Y, E, obs((*@\textcolor{blue}{P}@*)((*@\textcolor{red}{o2}@*)))) :- neg(X, Y, E, e_pred((*@\textcolor{blue}{P}@*))).

(*@\label{lstline:choice_rule}@*)1 { pos(X, Y, E, obs((*@\textcolor{blue}{P}@*)((*@\textcolor{red}{o1}@*))));
pos(X, Y, E, obs((*@\textcolor{blue}{P}@*)((*@\textcolor{red}{o2}@*)))) } 2 :-
    pos(X, Y, E, e_pred((*@\textcolor{blue}{P}@*))).

1 { neg(X, Y, E, obs((*@\textcolor{blue}{P}@*)((*@\textcolor{red}{o1}@*)))); 
neg(X, Y, E, obs((*@\textcolor{blue}{P}@*)((*@\textcolor{red}{o2}@*)))) } 2 :-
    neg(X, Y, E, a_pred((*@\textcolor{blue}{P}@*))).
\end{lstlisting}
\end{listing}

We then enforce the mutual exclusivity of each atomic formulae between edges starting from the same node and reaching different ones \cite{Furelos-Blanco_Law_Jonsson_Broda_Russo_2021}, using the $\texttt{neg}$ and $\texttt{pos}$ facts of the propositions, ground atoms and the mapped ground atoms associated with quantified atoms.

\begin{figure}[h]
    \centering
    \captionsetup{justification=centering}
    \includegraphics[width=\linewidth]{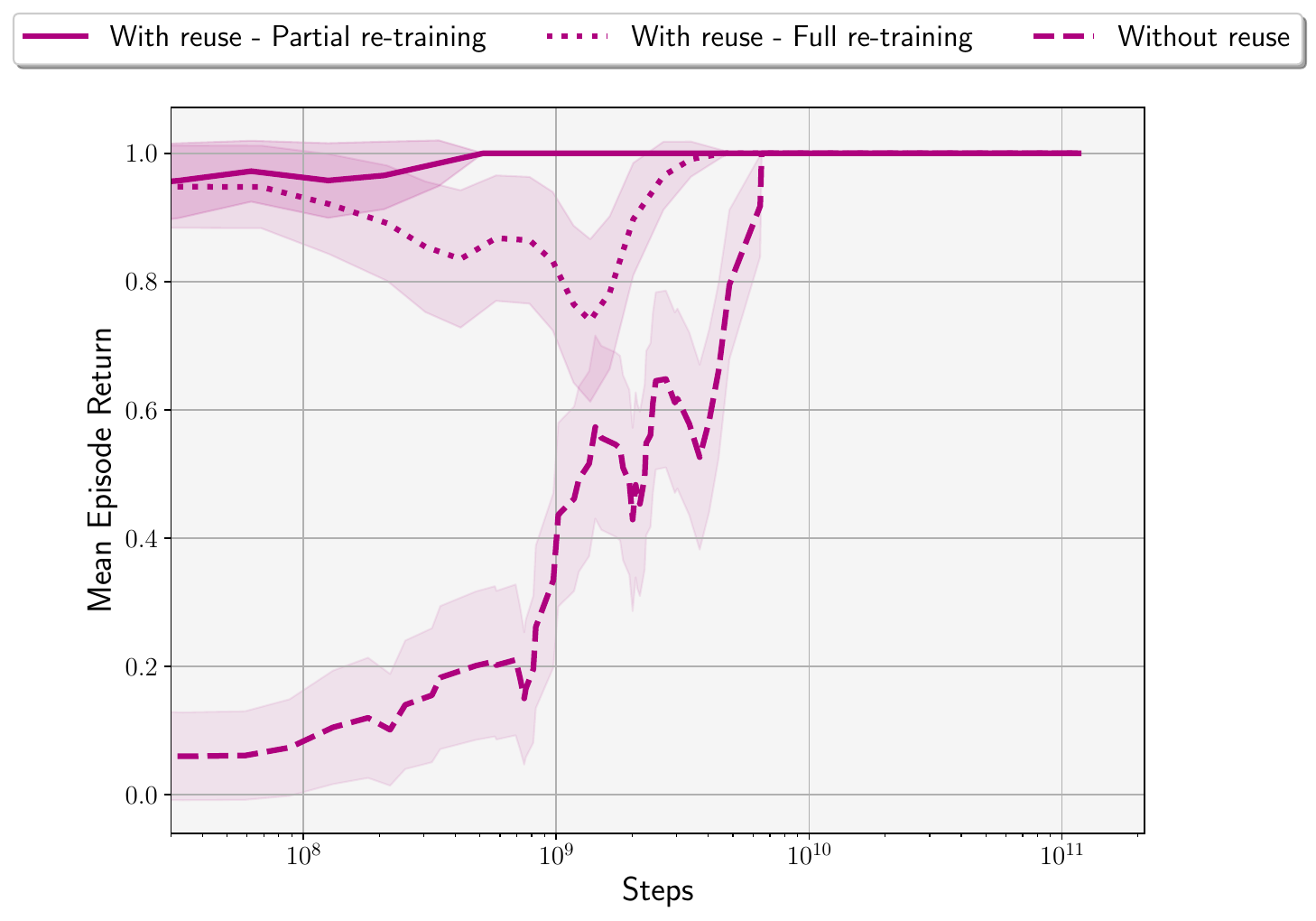}
    \caption{Transfer to the \textsc{4-Yellow} task.}
    \label{fig:transfer_4}
    \Description{Performance of transferring the ``2 Yellows'' policies to the ``4 Yellows'' task.}
\end{figure}

\begin{figure}[h]
    \centering
    \captionsetup{justification=centering}
    \includegraphics[width=\linewidth]{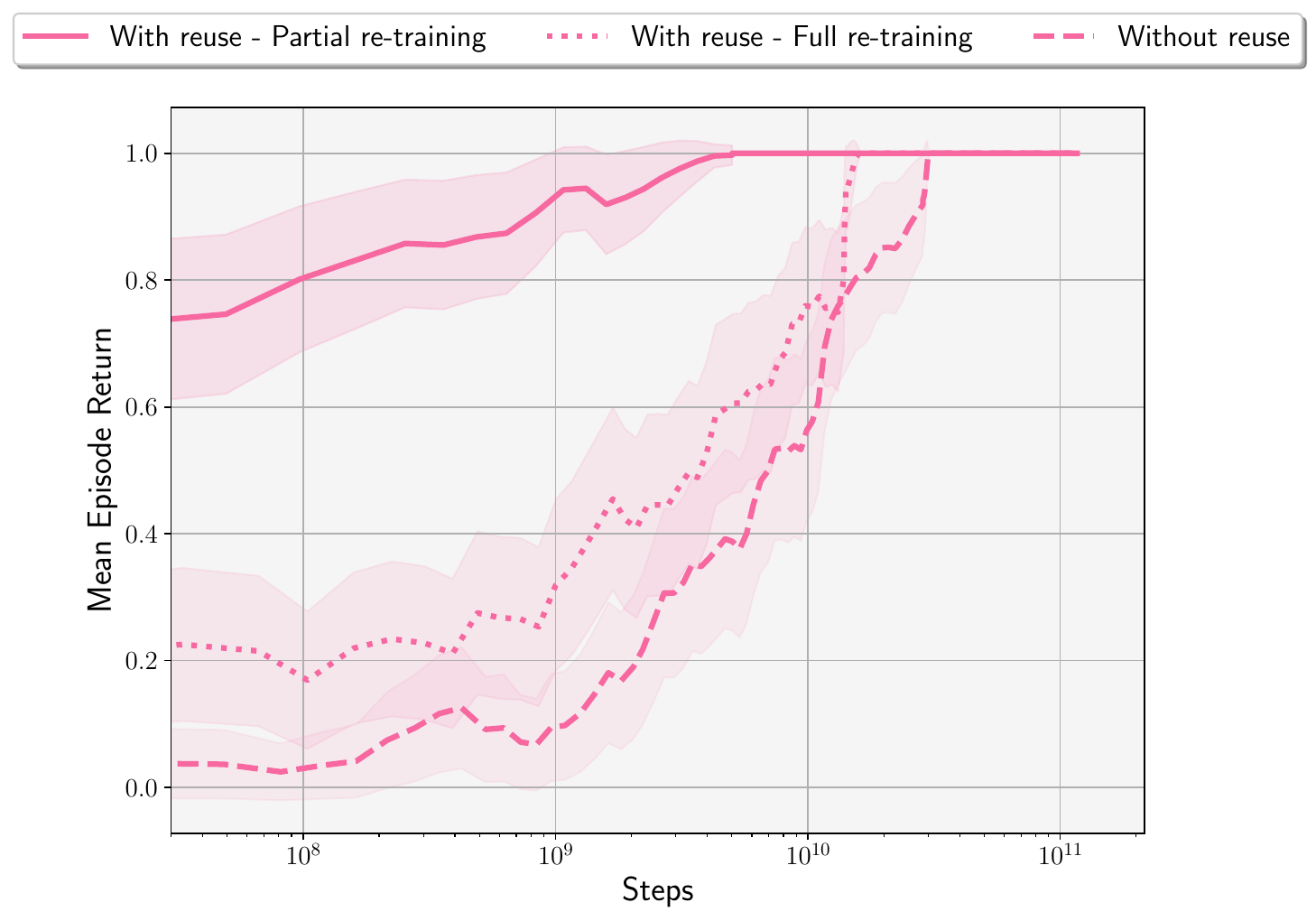}
    \caption{Transfer to the \textsc{6-Yellow} task.}
    \label{fig:transfer_6}
    \Description{Performance of transferring the ``2 Yellows'' policies to the ``6 Yellows'' task.}
\end{figure}

\section{Additional Transfer Experiments}\label{apdx:transfer}

We show the benefit of our multi-agent learning approach compared to a single policy, when transferring the RM and the policy to a new environment. The single policy is trained on the cross-product between the environment state space $\envStatesSet$ and the RM state space $\statesSet$. In the new environment, the number of yellow checkpoints ($\checkpointEx{yellow}{}$) is increased from \emph{two} to \emph{four} and \emph{six}, but the task remains the same: ``visit all yellow checkpoints $\checkpointEx{yellow}{}$, before going to the location $\goalCell$''. We observe that for the single policy, the transfer is ineffective and the policy does not succeed in the new environment. On the other hand, we observe that our multi-agent approach succeeds in the new environment.

We also analyse the impact of reusing policies that have been trained on a simpler task (\textsc{2-Yellow}) into more complex ones: \textsc{4-Yellow} (\figref{fig:transfer_4}) and \textsc{6-Yellow} (\figref{fig:transfer_6}). We examine different configurations and compare the number of iterations to converge with the base case where the policies are learnt from scratch \textit{(Without reuse)}. The first configuration \textit{(With reuse - Partial re-training)} corresponds to the case where we know `a priori' the policies that must be retrained. In that case, we can simply retrain a subset of the policies and use the others in inference mode during training. We observe a huge speed-up in the learning of the policies with the agents converging in fewer iterations for both \textsc{4-Yellow} and \textsc{6-Yellow}. In the second case, we retrain all the policies \textit{(With reuse - Full re-training)} but use the transferred policies as a warm start. In this configuration we observe a decent speed-up for the \textsc{4-Yellow} task but only 
a marginal improvement for \textsc{6-Yellow}.

\begin{figure}[h]
    \centering
    \captionsetup{justification=centering}
    \includegraphics[width=0.75\linewidth]{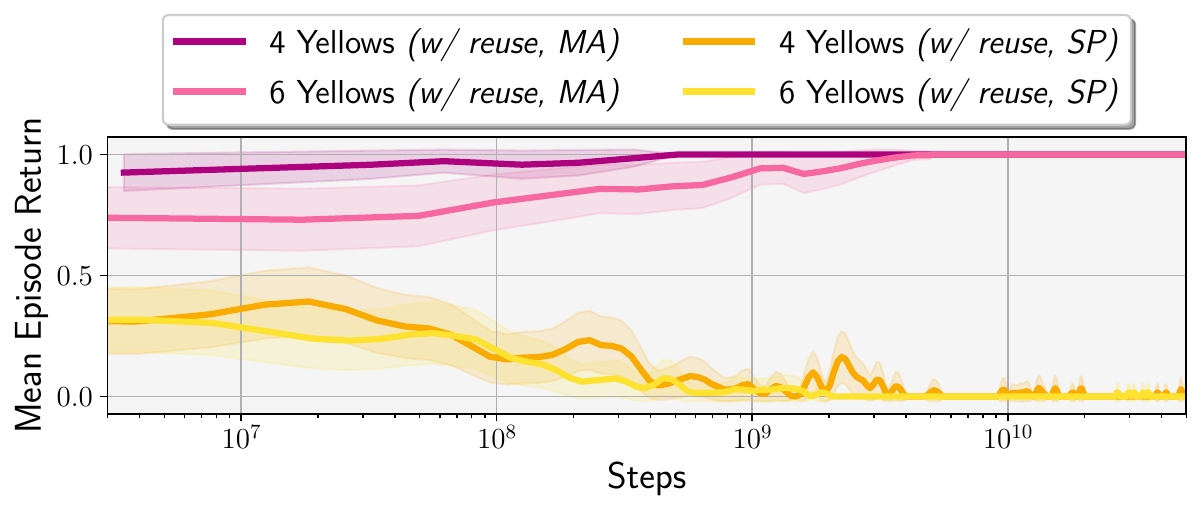}
    \caption{Transferring the RM and policies learnt with a single policy (\emph{SP}) versus our multi-agent approach (\emph{MA}).}
    \Description{}
    \label{fig:crm_transfer}
\end{figure}

\update{\figref{fig:crm_transfer} shows the benefit of our multi-agent approach by comparing the performance of transferring the \FORM{} learnt with a single policy with the performance of transferring the \FORM{} learnt with out multi-agent approach. By having separate policies, the adaptation to a new environment configuration is feasible. While the single policy associated with the learnt \FORM{} fails to transfer to a different configuration, our multi-agent approach is able to re-train only the policies that need retraining but re-using the other policies.}

\end{document}